\documentclass[10pt,twocolumn,letterpaper]{article}

\usepackage[
    letterpaper,
    top=0.75in,
    bottom=0.75in,
    left=0.75in,
    right=0.75in,
    columnsep=0.25in
]{geometry}

\usepackage{amsmath,amssymb,amsfonts,amsthm}
\usepackage{algorithmic}
\usepackage{algorithm}
\usepackage{graphicx}
\usepackage{textcomp}
\usepackage{xcolor}
\usepackage{booktabs}
\usepackage{multirow}
\usepackage{bm}
\usepackage{float}
\usepackage{stfloats} 
\usepackage[nospread]{flushend} 
\usepackage{placeins} 
\usepackage{longtable,array,pdflscape}

\usepackage{adjustbox} 
\usepackage[subtle]{savetrees}


\definecolor{linkblue}{rgb}{0.21,0.49,0.74}
\usepackage[pagebackref,breaklinks,colorlinks,allcolors=linkblue]{hyperref}

\definecolor{darkgreen}{RGB}{0,100,0}
\definecolor{darkred}{RGB}{139,0,0}

\newcommand{\poscell}[2]{\textcolor{darkgreen}{\pmcell{#1}{#2}}}
\newcommand{\negcell}[2]{\textcolor{darkred}{\pmcell{#1}{#2}}}
\newcommand{\posdiff}[1]{\textcolor{darkgreen}{\ensuremath{#1}}}
\newcommand{\negdiff}[1]{\textcolor{darkred}{\ensuremath{#1}}}

\newcommand{\R}{\mathbb{R}}
\newcommand{\vz}{\mathbf{z}}
\newcommand{\vx}{\mathbf{x}}
\newcommand{\ve}{\mathbf{e}}
\newcommand{\vP}{\mathbf{P}}
\newcommand{\second}[1]{\underline{#1}}
\newcommand{\pmcell}[2]{#1{\scalebox{0.68}{\,($\pm$#2)}}}

\title{CalibTrack: Exploring Learned Heteroscedastic Covariance\\
for Robust Multi-Camera Pedestrian Tracking}
\date{}

\author{Jack Roberts$^{*}$ \quad
Aditya Iyer$^{*}$ \quad
Nora Ayanian\\[0.35em]
\small\texttt{\{aditya\_iyer, jack\_roberts2, nora\_ayanian\}@brown.edu}\\[-0.15em]
\small $^*$Equal contribution
}

\begin{document}
\maketitle

\begin{abstract}
Analytical multi-view multi-object tracking (MV-MOT) systems rely on measurement covariance estimates to guide association, fusion, and state estimation. These covariance estimates are typically heuristic, requiring hand-tuning to the specific deployment scenario. We investigate how learned heteroscedastic measurement covariance affects calibration and robustness in analytical multi-camera tracking. To support this analysis, we introduce CalibTrack, a detector-side predictor that replaces uniform measurement covariance with per-detection covariance. We evaluate CalibTrack on the \textit{WildTrack} and \textit{MultiviewX} pedestrian tracking datasets with multiple analytical tracker cores, including an identity-informed GM-PHD, JIPDA, and pruned $\delta$-GLMB filter. We additionally evaluate robustness by simulating adverse weather conditions with stable diffusion, Gaussian noise, camera browning, and camera dropout. CalibTrack improves average adverse tracking performance over confidence-scaled covariance, while producing better calibrated predictions, indicating the value of heteroscedastic covariance within analytical tracking. Comparisons with stronger heuristic covariance methods show that the benefits of learned covariance remain bounded by the quality of available observations and the downstream tracker choice. Together, these findings provide a principled exploration of learned covariance estimation and its potential for uncertainty-aware analytical tracking.

\end{abstract}

\begin{figure*}[!tbp]
\centering
\includegraphics[width=\textwidth, trim={0in 2in 0in .25in},clip]{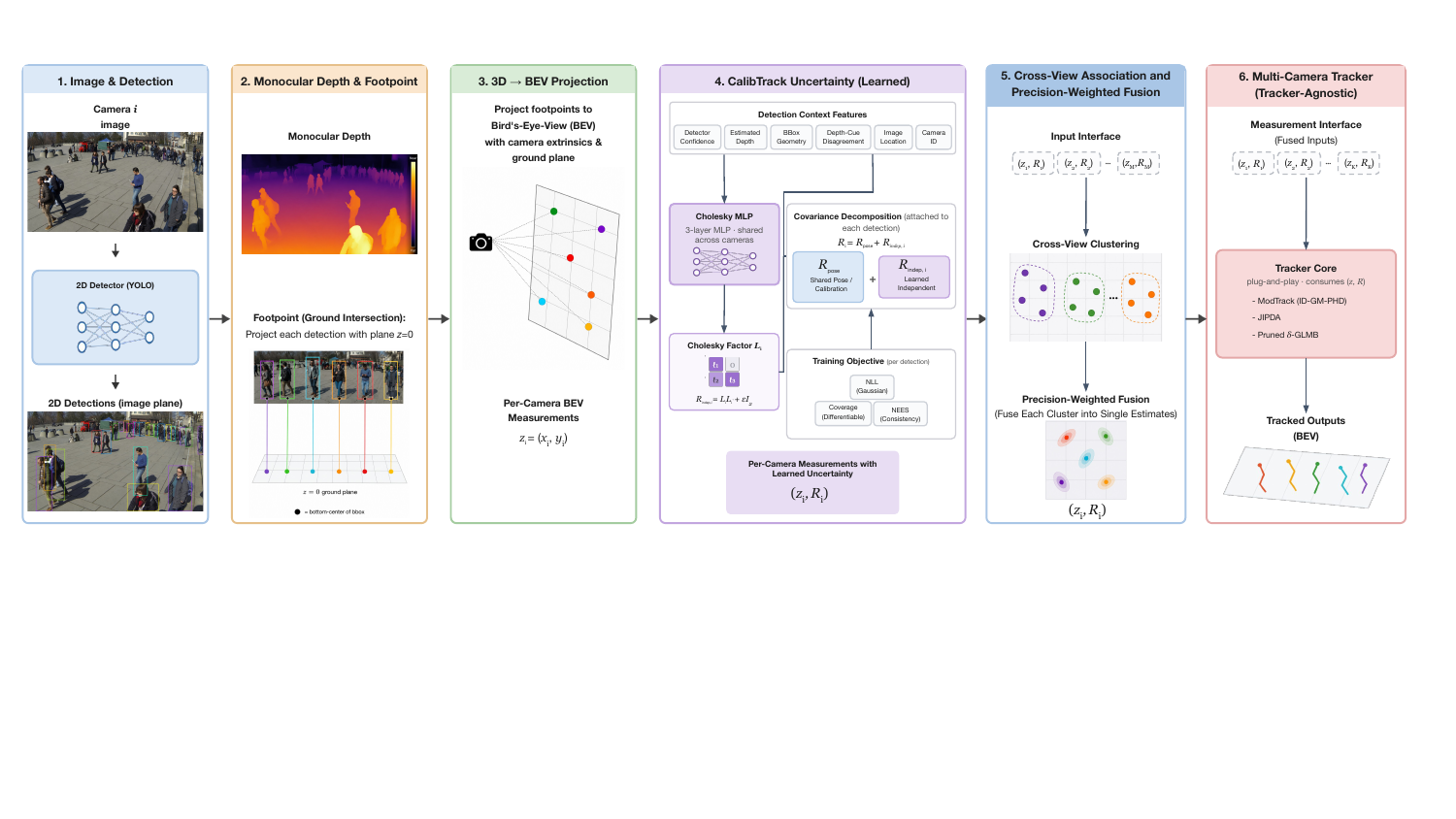}
\caption{\textbf{CalibTrack pipeline.} Per-camera detections (Stage~1) are projected to BEV via monocular depth and footpoint projection (Stages~2--3), giving position $\vz_i$. A Cholesky-parameterized MLP (Stage~4) predicts a context-conditioned independent covariance, combined with the fixed shared term. The measurement-covariance pairs are clustered and precision-fused (Stage~5), then consumed by an interchangeable analytical tracker core (Stage~6): GM-PHD, JIPDA, or pruned $\delta$-GLMB. CalibTrack modifies only Stage~4. All other stages are inherited from ModTrack.}
\label{fig:overview}
\end{figure*}

\section{Introduction}
\label{sec:intro}

Multi-camera pedestrian tracking is a critical task for surveillance, autonomous-vehicle perception, and embodied agents. For years, classical analytical filters~\cite{vo2006gmphd, musicki2004jipda, reuter2014lmb, vo2014delta_glmb} have been applied to this task due to their principled probabilistic treatment of multi-object state estimation, uncertainty, and data association. This uncertainty is represented through explicit covariance estimates which play a central role throughout the tracking pipeline.

These classical filters assume known noise covariance matrices, however, in practice these matrices are partially or fully unknown and may vary over time. Existing adaptive and learned filtering methods address this by estimating or refining covariance within the filtering process~\cite{mehra1970identification,mehra1972approaches,odelson2006new,huang2018novel,zhang2020fifty,dunik2017survey,haarnoja2016backprop,revach2022kalmannet,kloss2021train,mortada2025recursive,levy2025anukf,cohen2024akit}. However, in multi-camera tracking, covariance affects earlier decisions such as cross-view association and precision-weighted fusion before filtering begins. This motivates learning calibrated covariance at the detector–tracker interface, so that uncertainty is reliable throughout the full tracking pipeline rather than corrected only within the final state estimator.

Recent years have seen the rise of end-to-end learned tracking systems~\cite{ali2025directional, alturki2025attention,  mvtrajecter, tracktacular, mcblt, earlybird, atlasfusion} that process images directly into a shared bird’s-eye-view (BEV) representation and perform joint reasoning in feature space, eliminating the need for explicit measurement covariance information. These systems often achieve strong tracking performance but provide limited explicit uncertainty information for downstream reasoning, auditing, or decision-making.

ModTrack~\cite{iyer2025modtrack} bridged these two paradigms by combining learned perception with analytical tracking through an explicit detector-side $(\vz, R)$ representation. Rather than estimating and updating uncertainty inside the tracker, ModTrack's perception module emits a covariance ($R$) and BEV plane position ($\vz$) for every detection and propagates that uncertainty through association, fusion, and filtering. However, the covariance itself is generated by a hand-designed depth-based heuristic, limiting its ability to adapt to varying imaging conditions, viewpoints, and detector failure modes.

To study the limitations of hard-coded covariance, we introduce CalibTrack, a learned heteroscedastic covariance module that predicts the per-detection measurement covariance from detection context. CalibTrack learns detection-dependent covariances using a calibration-aware objective that combines Gaussian residual likelihood with consistency and coverage regularization, encouraging uncertainty estimates that are accurate, well-scaled, and reliable for downstream tracking. In contrast to adaptive and neural filtering approaches, CalibTrack performs detector-side covariance prediction rather than filter-side covariance refinement.  



To isolate the effect of covariance estimation, we hold perception outputs fixed and compare CalibTrack's learned $R$ with ModTrack's hand-tuned heuristic $R$, a confidence-scaled approach~\cite{liu2023confidence_mot}, and a dataset-level heuristic $R\times T$ global covariance recalibration baseline. We evaluate each covariance method with identity-informed GM-PHD, JIPDA, and pruned $\delta$-GLMB trackers on the WildTrack and MultiviewX pedestrian tracking datasets. Additionally, we simulate adverse conditions including diffusion-generated weather conditions, Gaussian pixel noise, camera browning, and camera dropout.

Across the evaluated adverse conditions, CalibTrack improves mean tracking performance over confidence scaling on both datasets and over scalar recalibration on WildTrack. Comparisons with ModTrack and with scalar recalibration on MultiviewX vary by tracker and metric. Together, these results show how learned, detection-dependent covariance can improve calibration and tracking and clarify when learning offers advantages over heuristic estimation.

The contributions of this work are three-fold.
\begin{itemize}
    \item \textbf{Controlled study of learned covariance.}
    We introduce CalibTrack and isolate the effect of its learned measurement covariance across multiple datasets and tracker cores, comparing it with heuristic, confidence-scaled, and global scalar-recalibrated covariance.

    \item \textbf{Analysis of calibration and tracking.}
    We investigate whether improved measurement calibration corresponds to improved downstream tracking and whether detection-dependent covariance improves calibration beyond global rescaling.

    \item \textbf{Robustness and limitations.}
    We simulate adverse conditions with stable diffusion and induce camera dropout to characterize where learned covariance helps. 
\end{itemize}

\section{Related Work}
\label{sec:related}

\textbf{Analytical multi-target tracking.}
Analytical multi-target trackers such as JIPDA, GM-PHD, and pruned $\delta$-GLMB maintain an explicit probabilistic state estimate and propagate uncertainty throughout the tracking pipeline. In these methods, measurement covariance plays a central role, defining Mahalanobis association gates, determining measurement likelihoods, and influencing how observations are fused and incorporated into state updates. Consequently, tracking performance depends not only on the accuracy of the measurements themselves, but also on the quality of the covariance attached to those measurements.

\textbf{Learnable covariance estimation.}
The true covariance statistics are often unknown or partially known and change over time as conditions evolve. To address this, classical adaptive filtering methods use innovation or residual statistics to update covariance estimates online, enabling the filter to adjust to changing noise characteristics during state estimation~\cite{mehra1970identification,mehra1972approaches,odelson2006new,huang2018novel,zhang2020fifty,dunik2017survey}. More recently, differentiable and neural filtering approaches have learned covariance, noise models, or Kalman-filter components directly from data~\cite{haarnoja2016backprop,revach2022kalmannet,kloss2021train,mortada2025recursive,levy2025anukf,cohen2024akit}. These methods focus on adapting uncertainty within a particular filtering framework, often using track history and filter dynamics as inputs. Consequently, they refine uncertainty after measurements have entered the tracker. In contrast, CalibTrack predicts covariance at the detector–tracker interface, allowing the same calibrated measurements to improve upstream association and fusion while remaining compatible with multiple analytical tracker cores.

\textbf{End-to-end learned MV-MOT tracking.}
A modern response to uncertain measurement models has been to eliminate them entirely. Recent multi-view tracking systems increasingly adopt end-to-end learned pipelines that operate directly on images rather than explicit probabilistic measurements~\cite{ali2025directional,alturki2025attention,mcblt,earlybird,atlasfusion,mvtrajecter,tracktacular, zhang2026mvtracktrans, sakai2026visibility}. These methods perform detection, association, and temporal reasoning within learned feature representations, often reasoning jointly across views in a shared latent BEV space. As a result, they avoid the need to specify or estimate measurement covariance explicitly, often achieving strong tracking performance through learned feature representations alone. However, the uncertainty associated with measurements is typically implicit within the network, making it difficult to inspect, calibrate, or propagate to downstream systems.

\textbf{Hybrid learned-analytical tracking.}
Recent work has explored a middle ground between fully analytical and fully end-to-end tracking systems by combining learned perception with analytical state estimation. ModTrack~\cite{iyer2025modtrack} adopts this approach through an explicit detector-side measurement interface, where each observation is represented by a BEV position $\vz$ and an associated measurement covariance $R$. This interface allows modern learned perception models to produce measurements while analytical trackers continue to perform association, fusion, and state estimation using explicit uncertainty. By propagating covariance throughout the tracking pipeline, such hybrid systems keep the interpretability and uncertainty-awareness of analytical methods while leveraging the representational power of learned perception. However, tracking performance depends on how accurately the measurement covariance captures localization errors, which can vary across detections and time. To better model this uncertainty, UncertaintyTrack~\cite{uncertaintytrack} incorporates detector localization uncertainty into image-plane multi-object tracking, while UncTrack~\cite{unctrack} uses localization uncertainty to assess prediction reliability and guide prototype memory updates in single-object tracking. CalibTrack focuses on learning per-detection BEV measurement covariance for cross-view association, precision-weighted fusion, and state estimation.

\textbf{MOT under adverse conditions.}
Single-view tracking performance is known to degrade under adverse atmospheric conditions~\cite{kirillova2024fogevaluatingrobustnessmultiple}. In contrast, robustness to environmental and sensor degradation remains largely unexplored in multi-view pedestrian tracking, where all dataset benchmarks are captured under clean conditions. To study this setting, we adopt a data augmentation approach introduced by Zhang et al.~\cite{bev-id} using Stable Diffusion~\cite{rombach2022highresolutionimagesynthesislatent} to augment WildTrack and MultiviewX in adverse weather conditions, enabling evaluation of tracking and covariance estimation under distribution shift.


\section{Methodology}
\label{sec:method}
\subsection{The $(z,R)$ Representation}
\label{interface}

The measurement-covariance representation $(\vz,R)$ is the core tracker- and modality-agnostic input used throughout the CalibTrack pipeline. Phases 1 through 5 in Figure~\ref{fig:overview} depict the process by which each $(\vz,R)$ BEV tuple is produced. First, an object detector (e.g., YOLOv11~\cite{yolo11_ultralytics}) produces bounding boxes for each pedestrian in each camera plane. Depth is a precision-weighted combination of ground-plane 'footpoint' backprojection and a bounding box height prior moderated by a learned depth from the ``Lift'' head of Lift-Splat-Shoot~\cite{philion2020liftsplatshootencoding}. We obtain a BEV coordinate $\vz$ for each detection by projecting the bounding box ``footpoint'' using the fused depth estimate. We clarify ``footpoint'' as the lowest central point of the bounding box.

Following ModTrack, each measurement covariance is decomposed as
\begin{equation}
R_i = R_{\mathrm{pose}} + R_{\mathrm{indep},i},
\end{equation}
where $R_{\mathrm{pose}}$ is the shared pose/calibration uncertainty and $R_{\mathrm{indep},i}$ is the camera-specific independent measurement uncertainty. We keep $R_{\mathrm{pose}}$ fixed because it models system-level extrinsic and calibration uncertainty shared across all detections. Treating it as detection-specific would double-count shared uncertainty during precision fusion. For each camera observation, CalibTrack predicts $\hat R_{\mathrm{indep},i}$, which is combined with $R_{\mathrm{pose}}$ to produce $R_i$. The resulting $(\vz_i,R_i)$ pairs are then clustered and fused to produce a unified set of measurements.

This covariance drives every downstream stage of the tracking pipeline. For detections $i,j$ from distinct cameras, cross-view association uses the squared Mahalanobis distance
$d_{ij}^2 = (\vz_i - \vz_j)^\top (R_i + R_j)^{-1}(\vz_i - \vz_j),$
independent components are precision-fused,
\begin{equation}
\vP_\text{indep}^{-1} = \sum_m R_{\text{indep},m}^{-1},
\qquad
\vP_\text{fused} = \vP_\text{indep} + R_\text{pose},
\label{eq:fusion}
\end{equation}
Equation~\ref{eq:fusion} assumes the camera-specific residuals represented by $R_{\mathrm{indep},m}$ are conditionally independent.

The Kalman update uses $S = HPH^\top + R.$
Consequently, changing $R$ affects association, fusion, and filtering even when the perception front-end and tracker are otherwise unchanged.

\subsection{Independent Covariance Head}

To predict the independent covariance term for a given camera detection, CalibTrack uses a 3-layer MLP. The module takes as input ten camera features (e.g., estimated depth, bounding-box geometry, detector confidence, depth-cue disagreement, image coordinates, and camera identity): $h_\phi : \R^{10} \to \R^3$.

Including camera identity helps model camera-specific effects, but leaves open whether the head learns transferable uncertainty cues or relies on camera-specific offsets. We therefore retrain the head with this identifier removed while holding perception and tracker settings fixed. We observe across ten training seeds that removing this input produces small changes in mean tracking scores (Supplementary (Supp.) Section~\ref{sec:camera_id_ablation}). Therefore, we keep the full ten-feature input throughout the main experiments.
This follows the broader heteroscedastic regression view that uncertainty should be conditioned on input context rather than a single global covariance~\cite{nix1994estimating,vega_brown2013cello,7353798,skafte2019reliable,stirn2023faithful}. Predicting covariance and measurements separately avoids the covariance head changing the detection mean, a failure mode in unconstrained heteroscedastic likelihood training~\cite{seitzer2022pitfalls}.

The MLP output parametrizes a lower-triangular Cholesky factor to ensure the output is a valid (symmetric positive definite) covariance matrix,
\begin{equation}
\begin{aligned}
L(\ell) &=
\begin{bmatrix}
e^{\ell_1} & 0 \\
\ell_2 & e^{\ell_3}
\end{bmatrix}, \\
\hat R_{\mathrm{indep},i} &= L(\ell_i)L(\ell_i)^\top + \epsilon I_2, \\
\hat R_i &= R_{\mathrm{pose}} + \hat R_{\mathrm{indep},i}.
\end{aligned}
\label{eq:cholesky}
\end{equation}
The exponential diagonal guarantees positive definiteness, while the off-diagonal term allows anisotropic uncertainty.

\subsection{Calibrated Covariance Training}
Training supervision is constructed from residuals between detected BEV points and their nearest ground-truth points, using a 3m Euclidean threshold rather than the tracker’s clustering and assignment logic to avoid circular supervision.

For detection $i \in \mathcal{D}$, let
$\ve_i = \vz_i - \vz_i^\star,$ $
s_i = \ve_i^\top \hat R_i^{-1}\ve_i,$
where $\vz_i$ is the projected measurement, $\vz_i^\star$ is the matched ground-truth BEV position, and $s_i$ is the quadratic Mahalanobis score.
The base objective is Gaussian negative log-likelihood,
\begin{equation}
\mathcal{L}_{\mathrm{NLL}}(\phi) =
\frac{1}{|\mathcal{D}|}\sum_i
\left[
\frac{1}{2}s_i(\phi)
+ \frac{1}{2}\log\det\hat R_i(\phi)
\right].
\label{eq:nll}
\end{equation}

NLL encourages covariance to explain the observed residuals, but it does not directly enforce calibration. A model can achieve a good average likelihood while remaining systematically overconfident in some conditions and underconfident in others. We therefore augment NLL with two calibration-oriented regularizers. The first targets covariance scale calibration through the normalized estimation error squared (NEES). The second targets probabilistic coverage, encouraging covariance ellipses to contain the expected fraction of residuals.

For a condition $c \in \mathcal{C}$ with samples $\mathcal{D}_c$, where $\mathcal{C}$ denotes the set of training conditions, define

$
\bar{s}_c = \frac{1}{|\mathcal{D}_c|}\sum_{i\in\mathcal{D}_c} s_i.$
For a calibrated 2D covariance, the quadratic score follows a $\chi^2_2$
distribution, implying $\mathbb{E}[s_i]=2$ and a normalized NEES of $1$.

The NEES regularizer therefore penalizes condition-wise deviations from this target,
\begin{equation}
\mathcal{L}_{\mathrm{NEES}} =
\frac{1}{|\mathcal{C}|}\sum_{c\in\mathcal{C}}(\bar{s}_c-2)^2.
\label{eq:nees_loss}
\end{equation}

While NEES constrains the average scale of the covariance, it does not guarantee that uncertainty ellipses contain the correct fraction of residuals. We therefore match empirical and nominal coverage probabilities.
$q_p = F^{-1}_{\chi^2_2}(p) = -2\log(1-p),$
so ideal empirical coverage satisfies $\Pr[s_i\le q_p]=p$.
Because the hard indicator is not differentiable, we use a soft coverage estimate
$\widehat{\mathrm{cov}}_{c,p}
=
\frac{1}{|\mathcal{D}_c|}
\sum_{i\in\mathcal{D}_c}
\sigma\!\left(\frac{q_p-s_i}{\tau}\right),$
where $\sigma(\cdot)$ is the logistic sigmoid and $\tau=0.5$ controls boundary softness.
Given $\mathcal{P} = \{0.8646647168,\,0.95\}$, the coverage loss is
\begin{equation}
\begin{aligned}
\mathcal{L}_{\mathrm{cov}} &=
\frac{1}{|\mathcal{C}||\mathcal{P}|}
\sum_{c\in\mathcal{C}}\sum_{p\in\mathcal{P}}
\left(\widehat{\mathrm{cov}}_{c,p}-p\right)^2.
\end{aligned}
\label{eq:coverage_loss}
\end{equation}

The first target corresponds exactly to $q_p=4$, the standard 2D ``$2\sigma$'' ellipse. The second probes a wider $95\%$ ellipse.

The final training objective is
\begin{equation}
\begin{aligned}
\mathcal{L} &=
\mathcal{L}_{\mathrm{NLL}}
+ \lambda_{\mathrm{NEES}}\mathcal{L}_{\mathrm{NEES}}
+ \lambda_{\mathrm{cov}}\mathcal{L}_{\mathrm{cov}}.
\end{aligned}
\label{eq:calibrated_loss}
\end{equation} 

For each dataset, we train the covariance head on pooled supervision from clean, diffusion-weather, Gaussian-noise, and browning conditions.
The same mixed regularization setting, $\lambda_{\mathrm{NEES}}=0.25$ and $\lambda_{\mathrm{cov}}=0.10$, is used for both WildTrack and MultiviewX. At evaluation time, the visual condition selects the condition-matched detector and depth model, but the covariance head remains a single mixed model for that dataset so the same covariance head is used across all evaluated conditions within each dataset.
Supp. Table~\ref{tab:loss_composition_conditions} reports a loss-composition ablation on the WildTrack dataset to empirically validate the three loss terms for calibrated measurement covariance estimation.

\subsection{Comparative Methods}

\subsubsection{Covariance Methods}

To compare CalibTrack's learned covariance framework, we consider three covariance baselines. The first is the heuristic covariance used by ModTrack (MT), \begin{equation}
R_{\mathrm{indep},i}^{\mathrm{MT}}
= \sigma_{d,i}^2\,\mathbf{j}_i\mathbf{j}_i^\top+\epsilon I_2,
\label{eq:modtrack_indep}
\end{equation}

where $\mathbf{j}_i$ is the local Jacobian mapping depth perturbations at the detected footpoint to a BEV position. The scalar $\sigma_{d,i}^2$ is derived from monocular depth estimates and increases with absolute depth, while additional inflation terms account for disagreement between multiple depth cues.

The second covariance method is a confidence-scaled approach inspired by Liu et al.~\cite{liu2023confidence_mot}, where \begin{equation}
R_i^{\mathrm{conf}} =
R_{\mathrm{pose}} +
\frac{R_{\mathrm{indep},i}^{\mathrm{base}}}{\max(c_i,\epsilon)}.
\label{eq:conf_scaled_indep}
\end{equation} 
The confidence scale is applied only to the independent term, not to $R_{\mathrm{pose}}$, in order to preserve ModTrack's shared-pose decomposition.

The third method applies a single scalar to every ModTrack covariance to isolate the effect of global covariance recalibration. The scalar is fitted on the mixed-supervision validation set. For each matched residual we compute the quadratic calibration score $s_i=e_i^\top (R_i^{\mathrm{MT}})^{-1} e_i$. For calibrated 2D Gaussian covariances these scores have expectation 2, so scaling every covariance by T scales every score by 1/T. Consequently, $T=\bar{s}/2$, where $\bar{s}$ denotes the average quadratic score over all matched residuals. To reduce sensitivity to extreme residuals, we compute both the ordinary mean and a 10\% trimmed mean of the quadratic scores. The trimmed estimate is used only when it differs from the ordinary estimate by more than 10\%. Otherwise the ordinary estimate is kept.
$
R_{\mathrm{pose}}^{T}
= T R_{\mathrm{pose}},$ 
$R_{\mathrm{indep},i}^{T}
= T R_{\mathrm{indep},i}^{\mathrm{MT}},$ 
$R_i^{T}
= T R_i^{\mathrm{MT}}.
$
Because both the independent camera terms and the shared pose term are scaled, precision fusion preserves the original relative weighting while producing exactly T times the original ModTrack fused covariance. The selected trimmed-mean scalars are T=0.1126 for WildTrack and T=0.1828 for MultiviewX.

\subsubsection{Tracker Cores}
To compare CalibTrack's transferability to various downstream trackers that consume the same fused $(z,R)$ outputs,  we consider three different tracker cores. The first tracker core is ModTrack's identity-informed GM-PHD tracker~\cite{iyer2025modtrack} for its competitive performance while propagating uncertainty. The second and third are the JIPDA~\cite{musicki2004jipda} and the pruned $\delta$-GLMB~\cite{vo2014delta_glmb} filters.
These backends are practical tracker cores, not claims of exact closed-form Bayes/RFS filtering: the perception front end has already clustered and fused camera detections, births are detection-driven because the datasets do not provide a separate birth model, and the association posteriors are ranked and pruned for tractability.
Thus, these rows test whether CalibTrack's Gaussian $(z,R)$ likelihood transfers across analytical association cores, not whether the implementation preserves the formal guarantees of an exact GM-PHD, JIPDA or pruned $\delta$-GLMB recursion.
We do not benchmark against end-to-end learned multi-view trackers in this controlled comparison because they do not expose the same covariance interface that allows isolation of the effect of $R$ on the detector side.

\subsection{Simulated Adverse Conditions}
\label{sec:protocol}

\begin{figure*}[!tbp]
\centering
\setlength{\tabcolsep}{2pt}
\resizebox{\textwidth}{!}{%
\begin{tabular}{ccccccc}
\textbf{Clean} & \textbf{Fog} & \textbf{Rain} & \textbf{Dust} & \textbf{Snow} & \textbf{Gauss} & \textbf{Brown} \\
\includegraphics[width=0.135\linewidth]{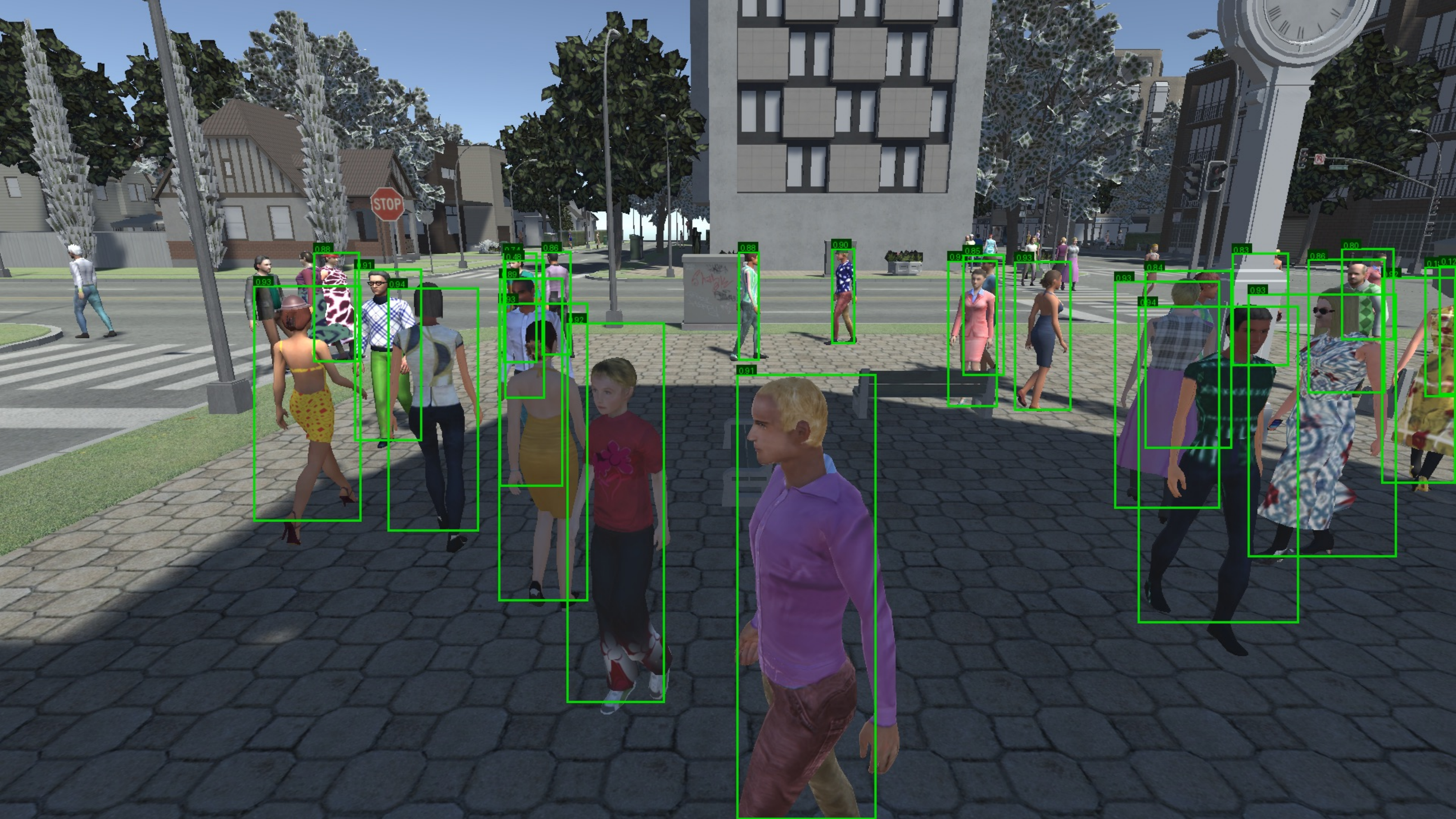} &
\includegraphics[width=0.135\linewidth]{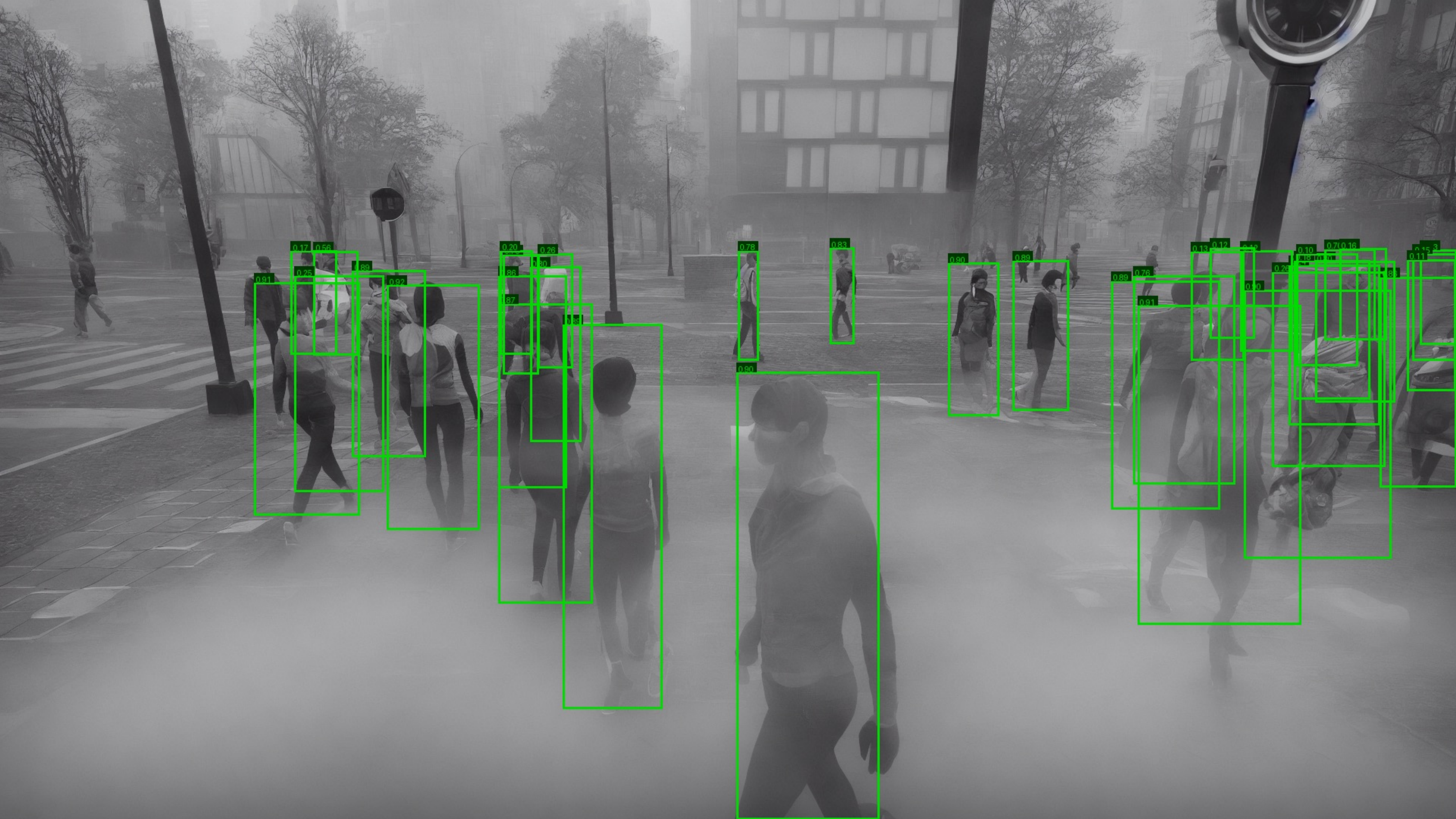} &
\includegraphics[width=0.135\linewidth]{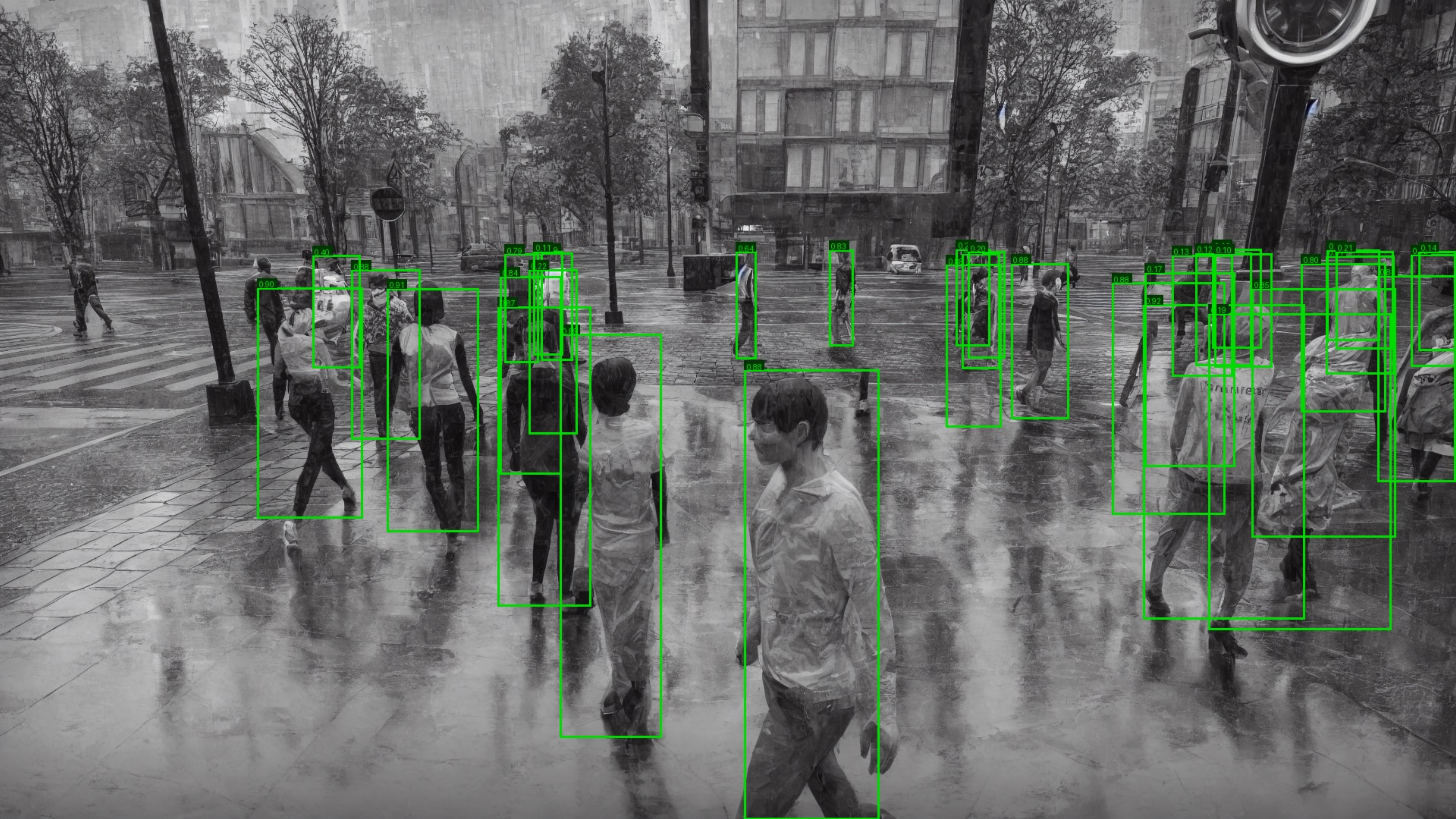} &
\includegraphics[width=0.135\linewidth]{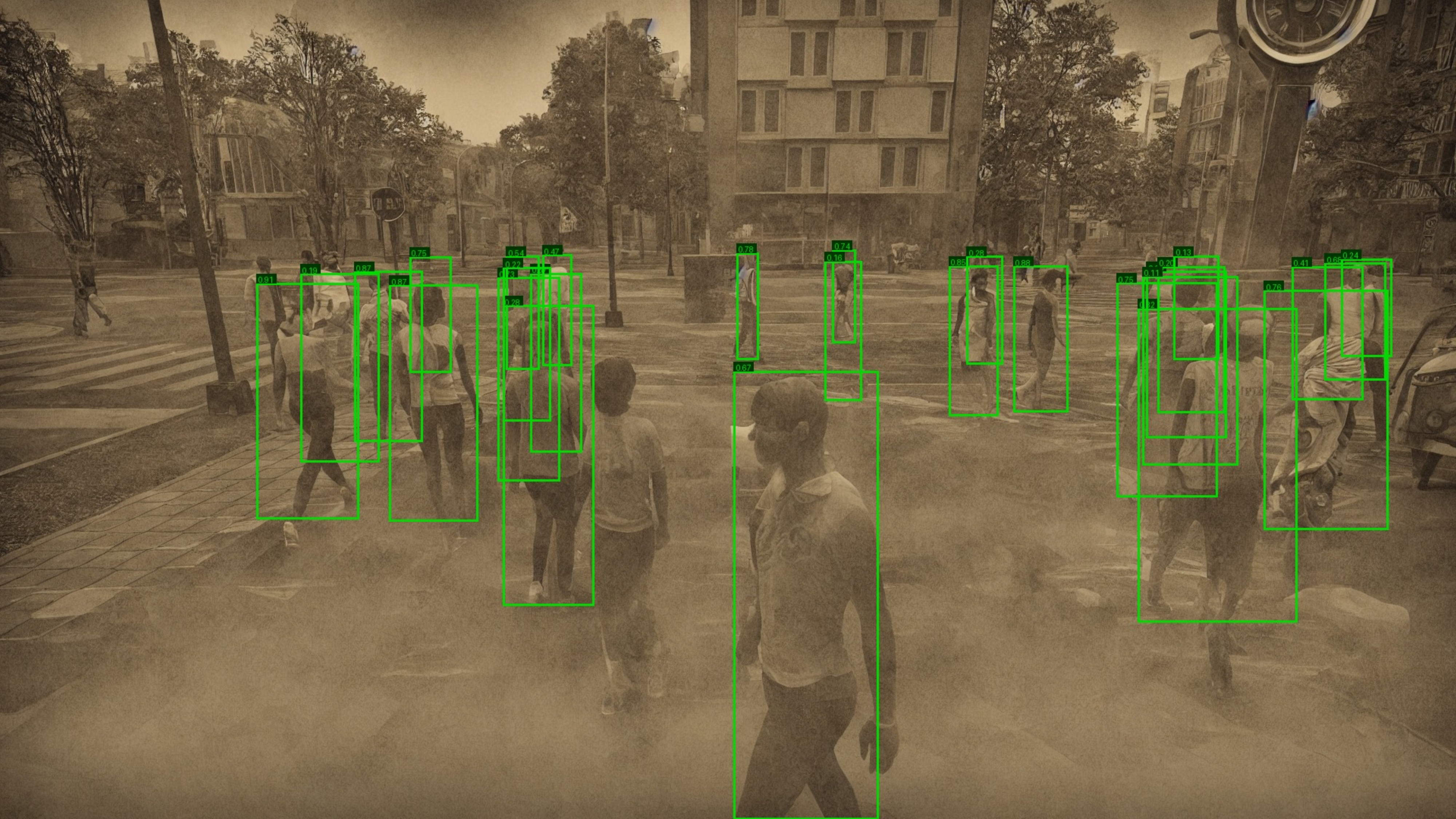} &
\includegraphics[width=0.135\linewidth]{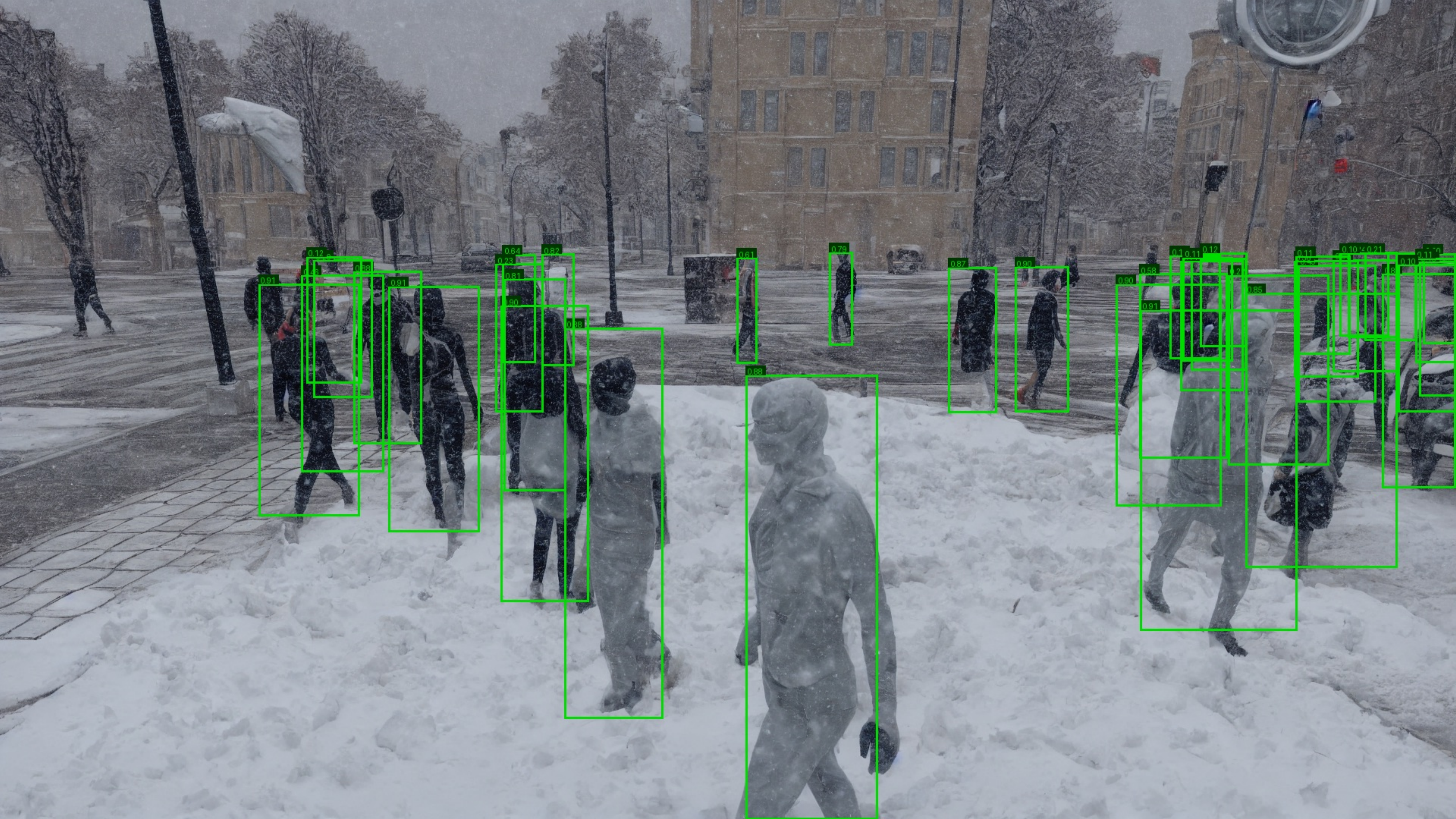} &
\includegraphics[width=0.135\linewidth]{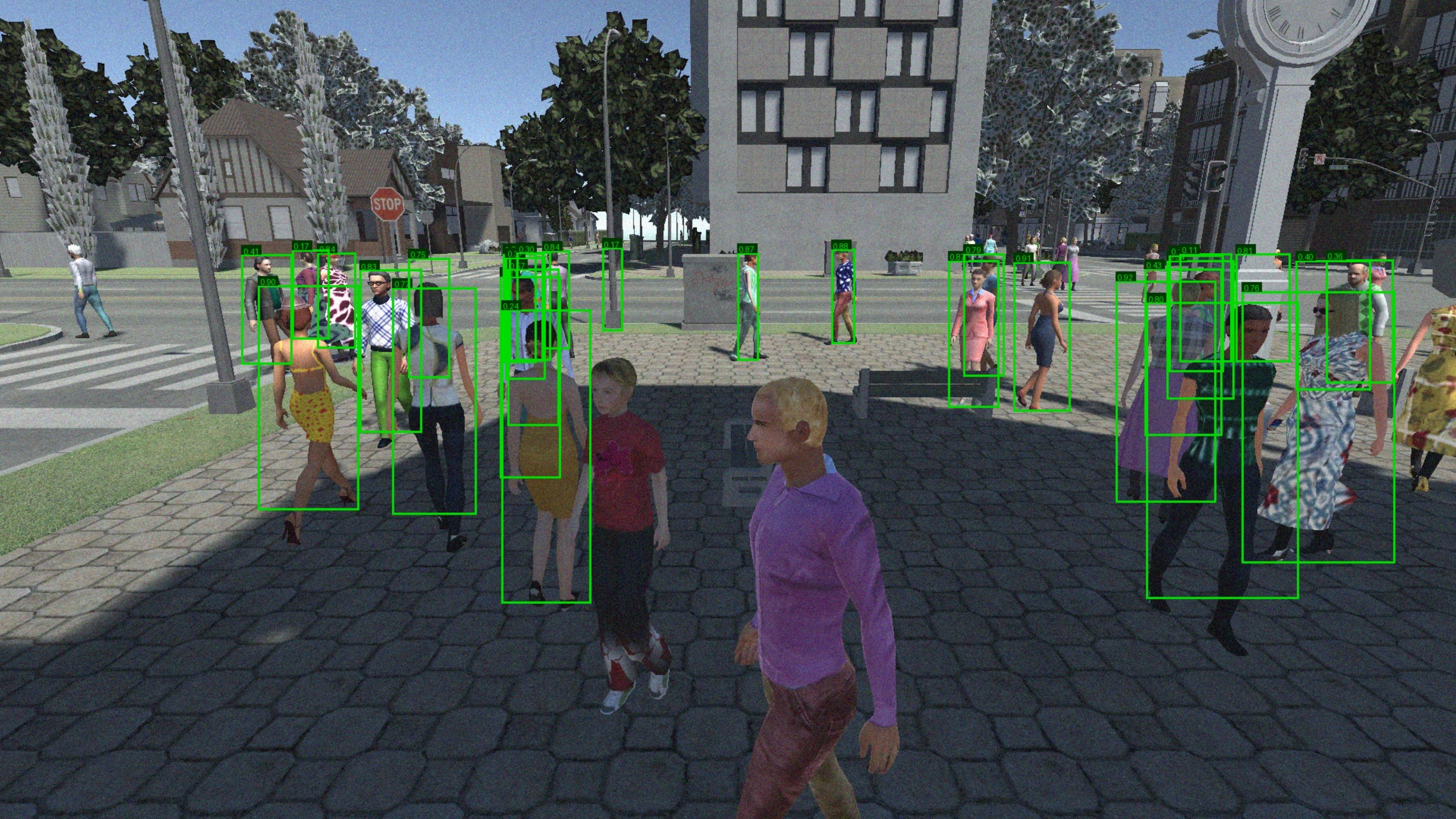} &
\includegraphics[width=0.135\linewidth]{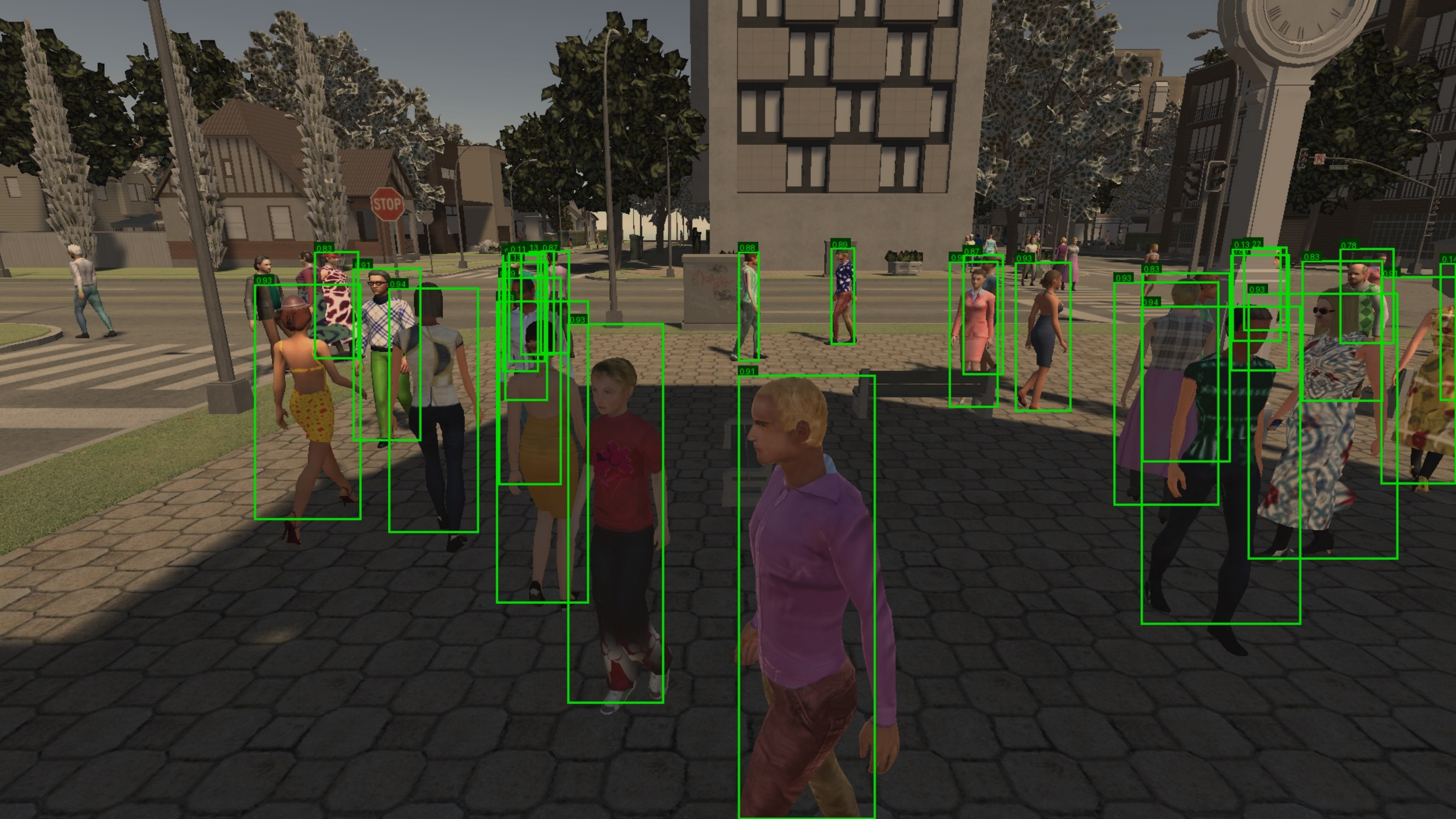} \\

\includegraphics[width=0.135\linewidth]{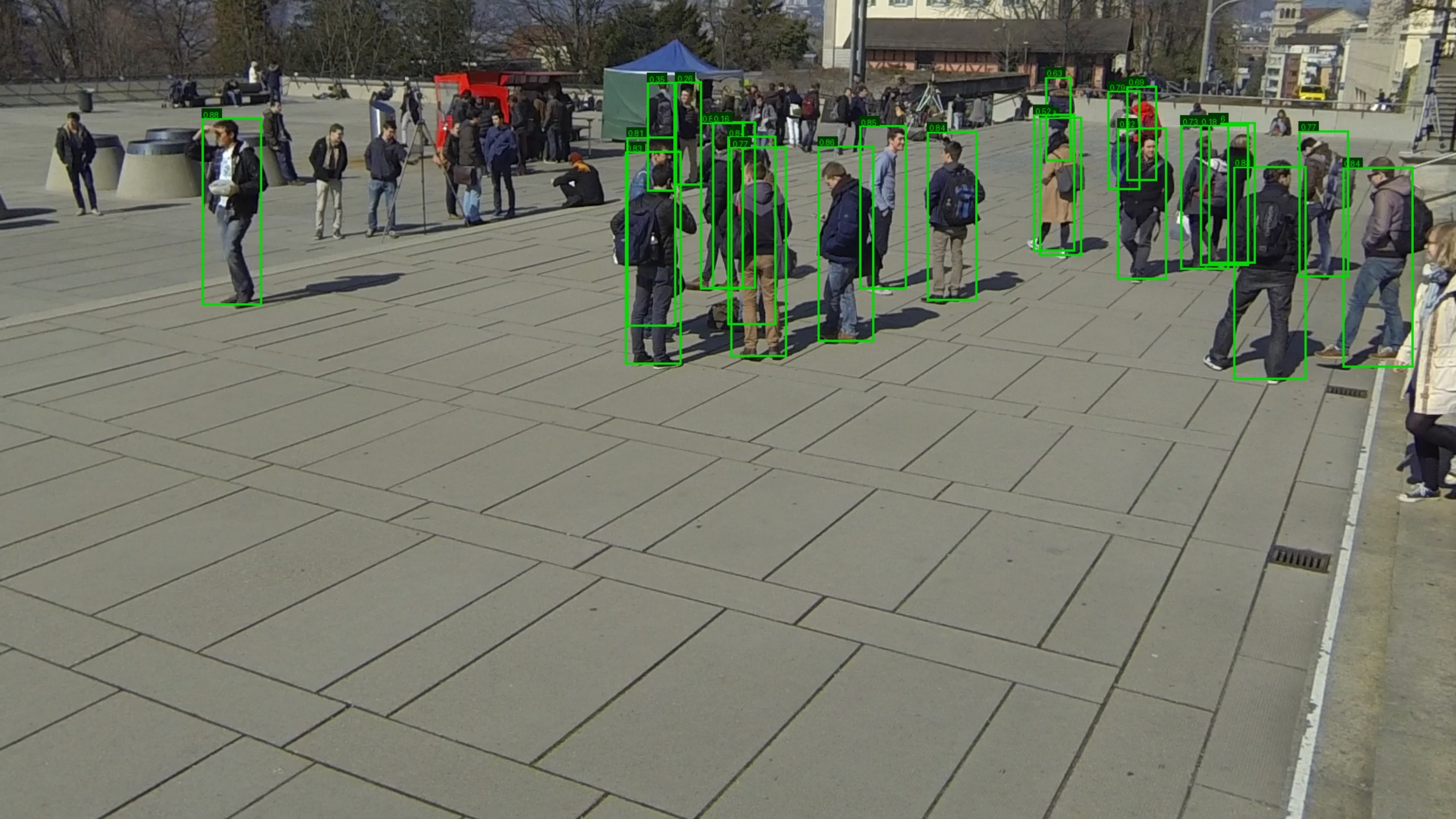} &
\includegraphics[width=0.135\linewidth]{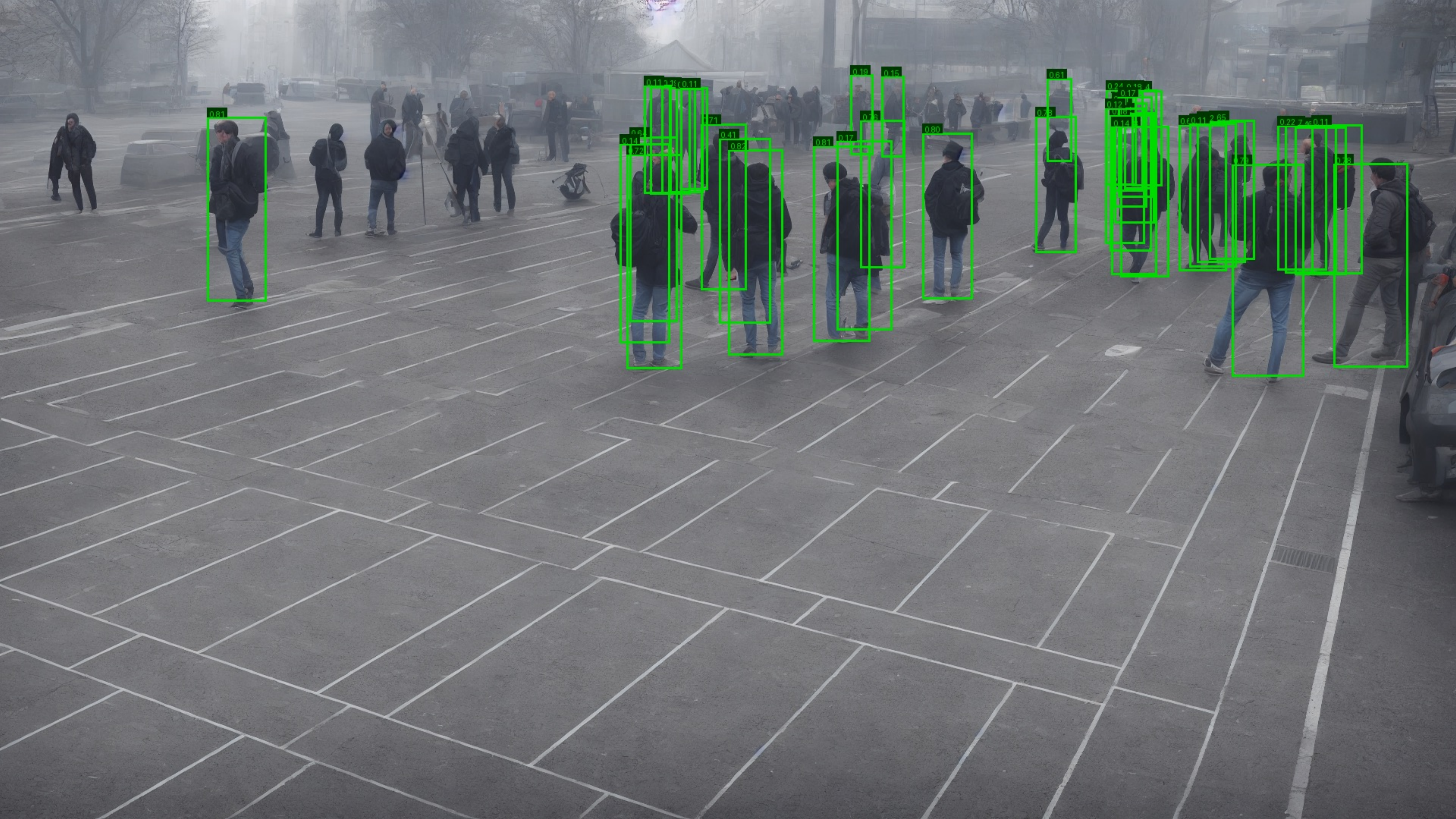} &
\includegraphics[width=0.135\linewidth]{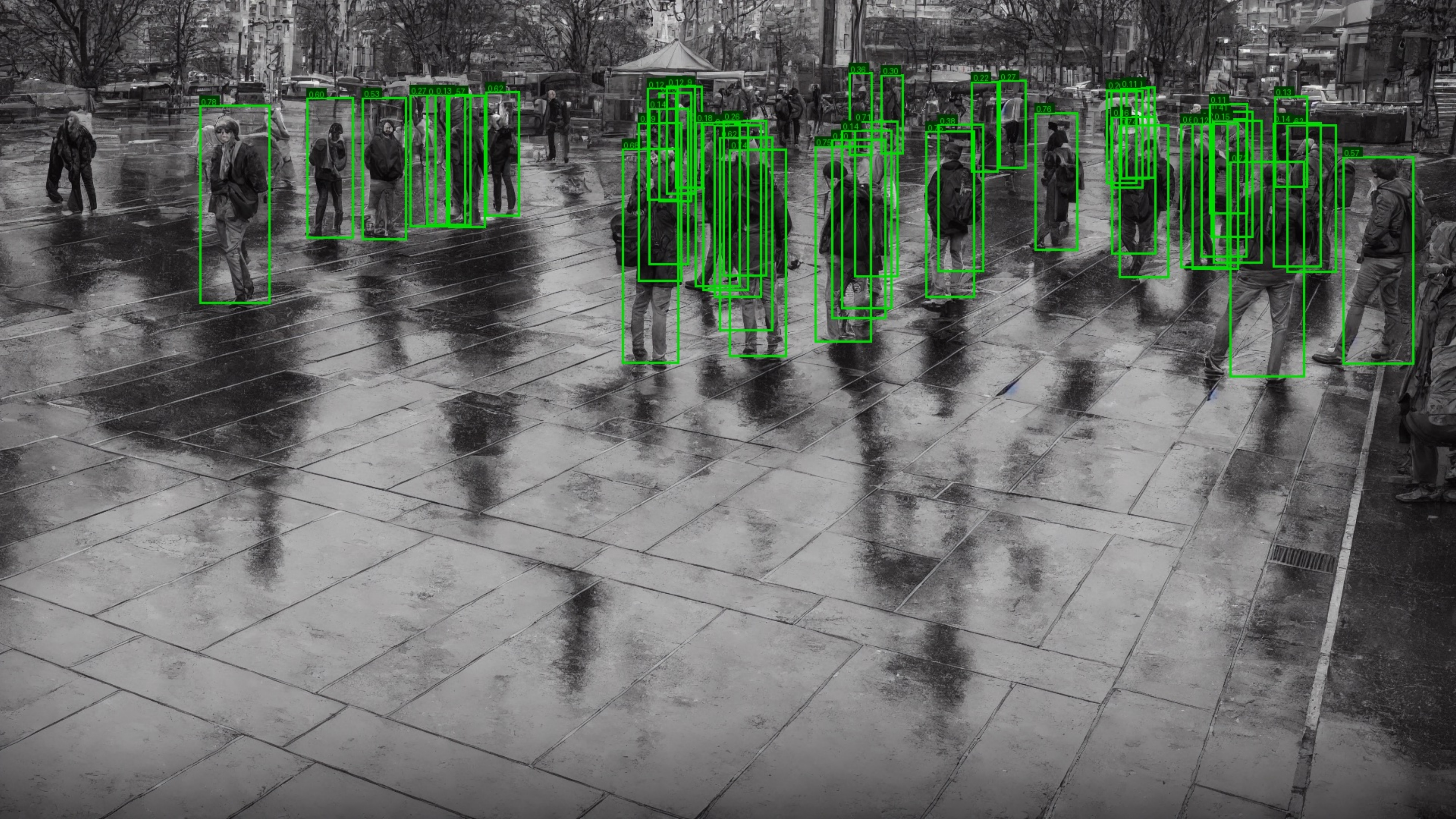} &
\includegraphics[width=0.135\linewidth]{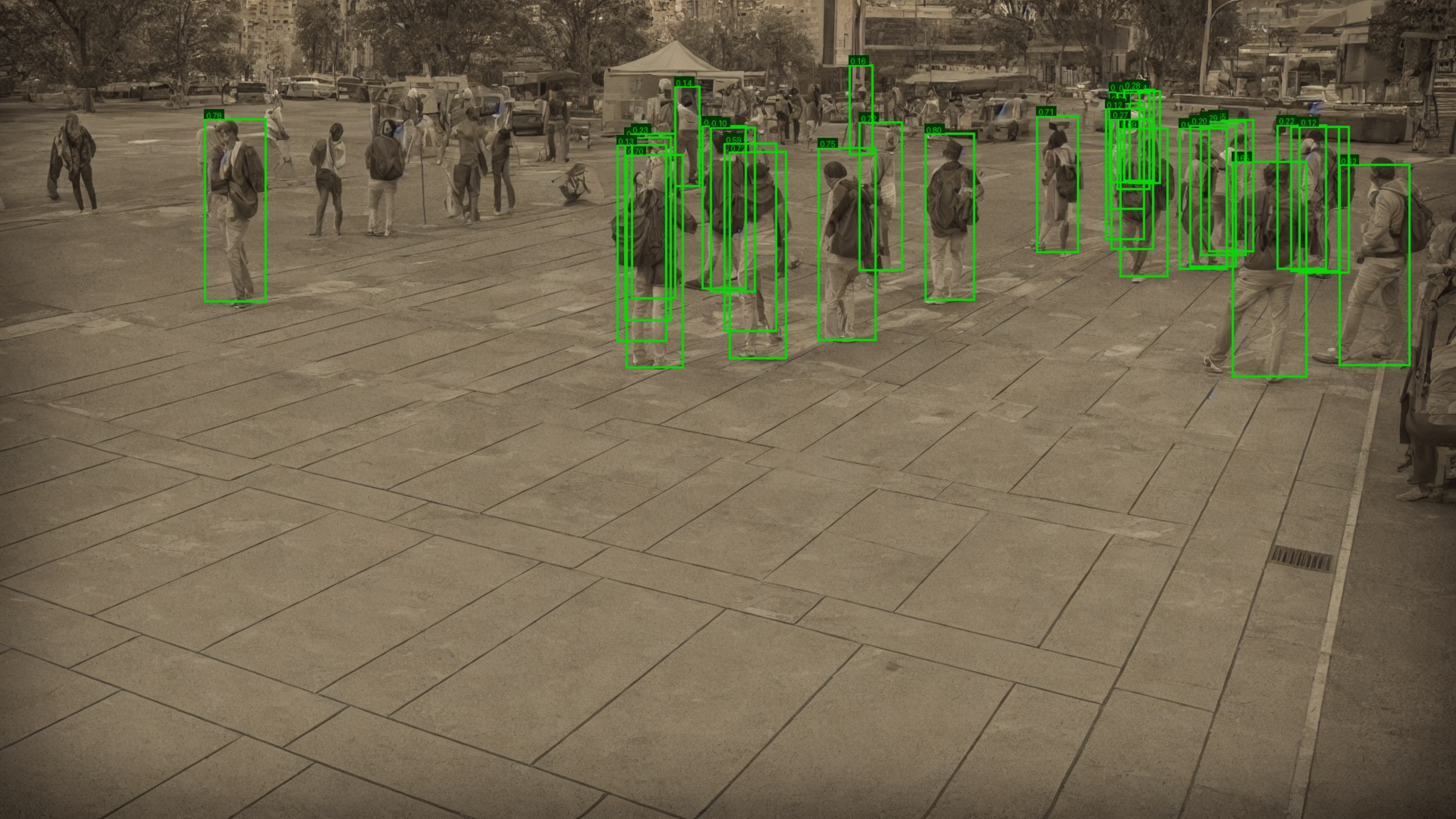} &
\includegraphics[width=0.135\linewidth]{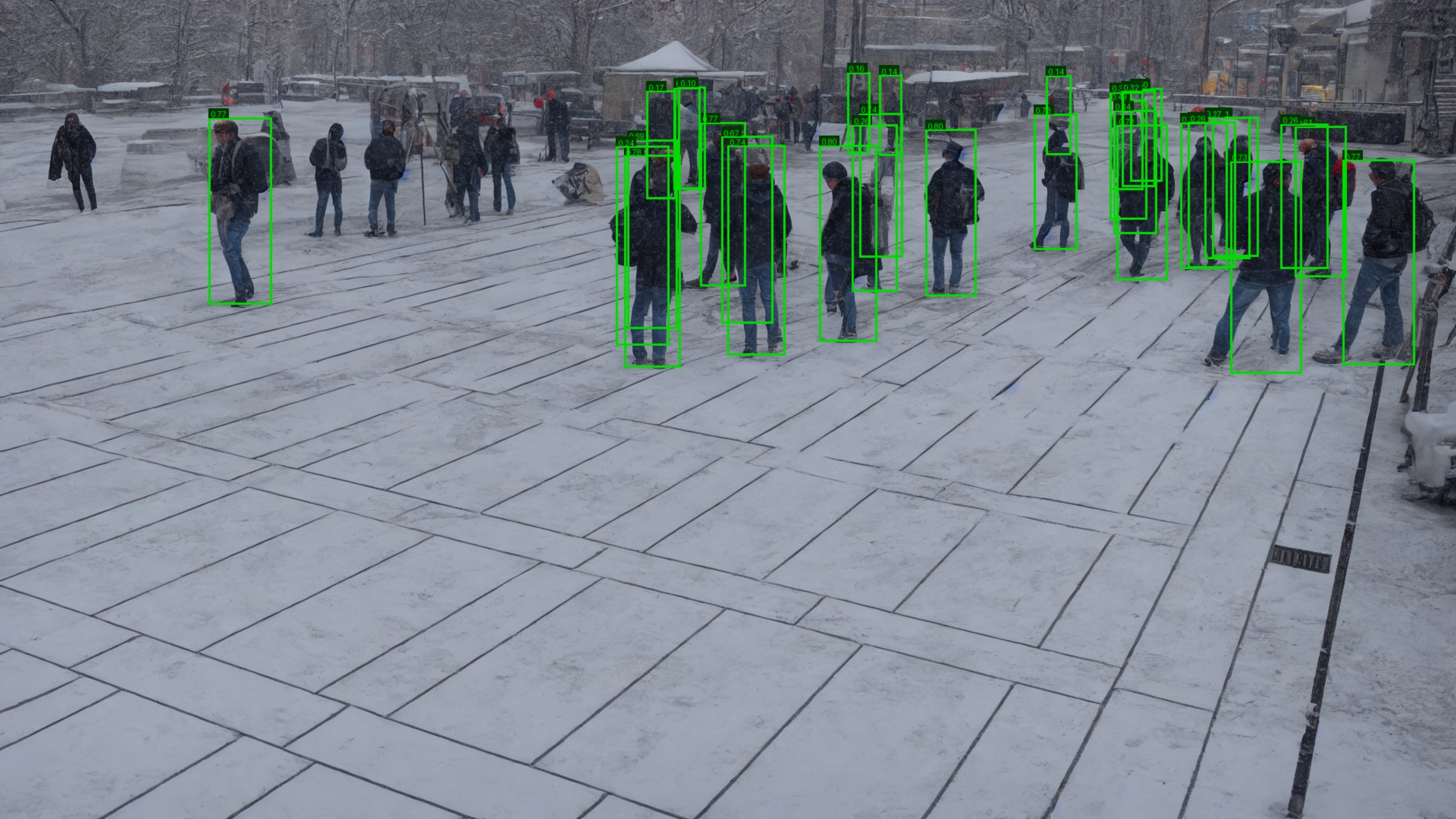} &
\includegraphics[width=0.135\linewidth]{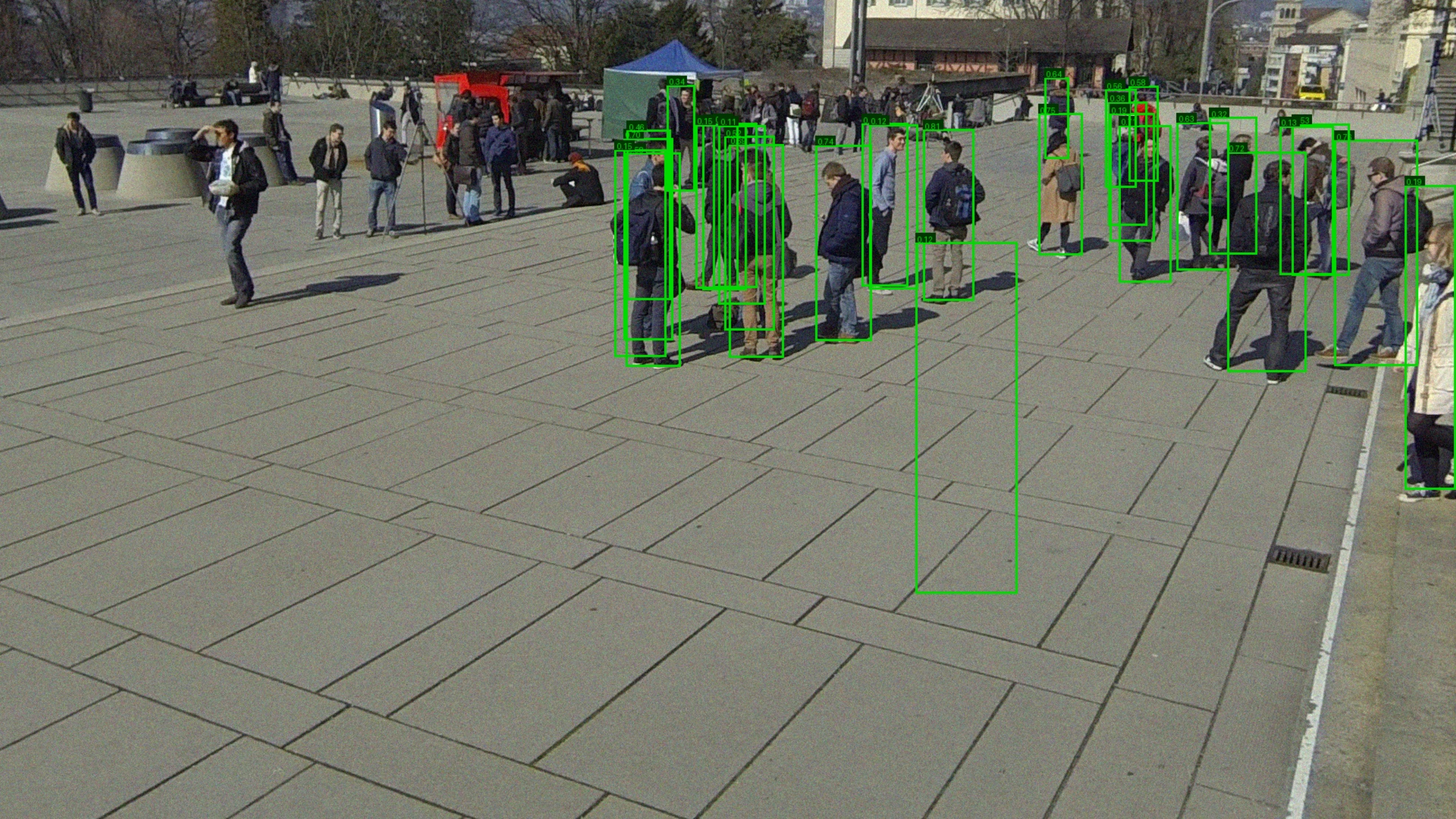} &
\includegraphics[width=0.135\linewidth]{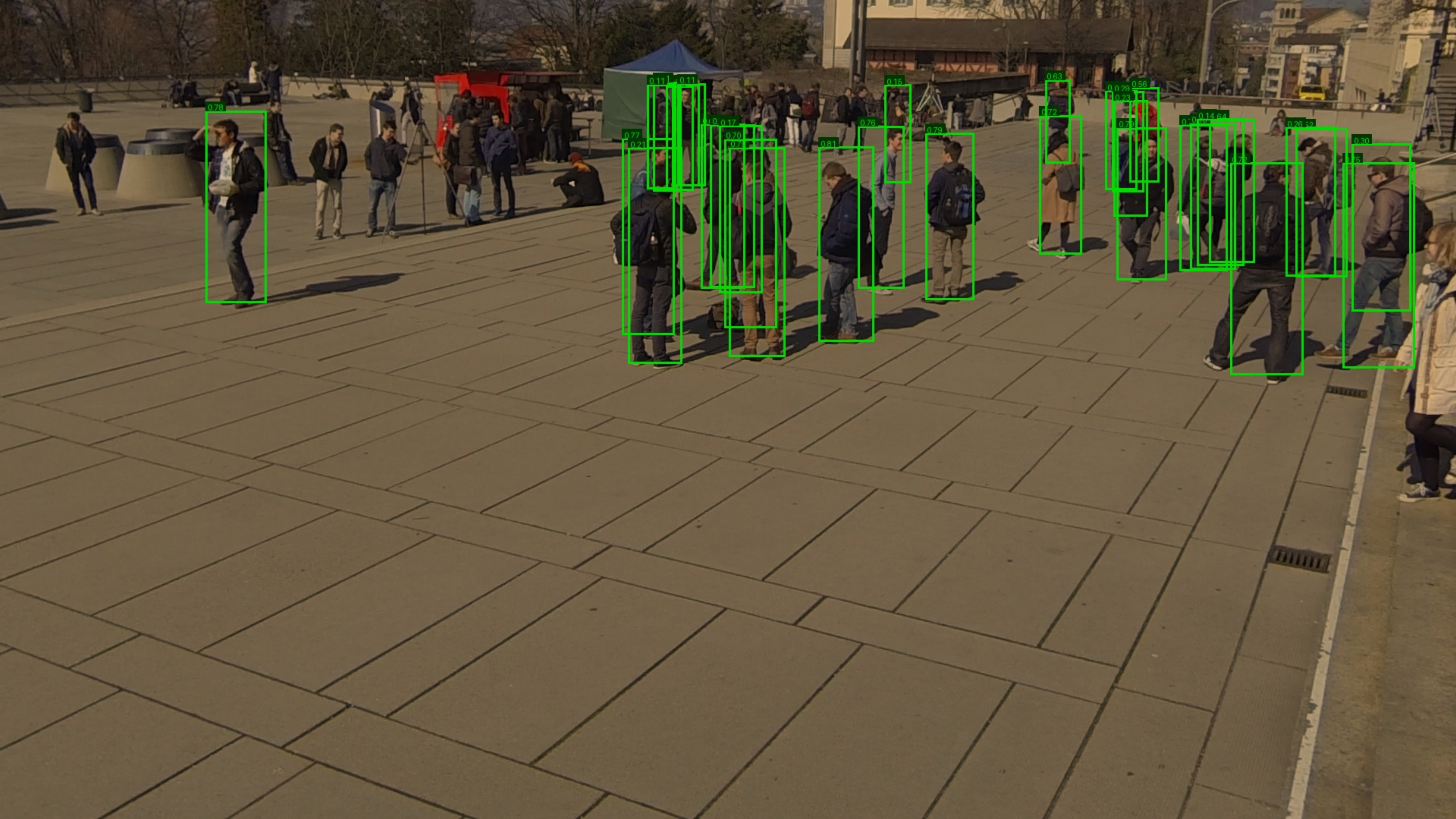} \\
\end{tabular}}

\caption{\textbf{Condition Visualization.} Detector outputs under clean, diffusion-weather, Gaussian-noise, and browning conditions. Boxes are produced by the condition-matched YOLO model. MultiviewX (top). WildTrack (bottom).}
\label{fig:weather_yolo}

\end{figure*}

\textbf{Weather conditions} on WildTrack and MultiviewX are produced by a data augmentation approach introduced by Zhang et al.~\cite{bev-id} using Stable Diffusion~\cite{rombach2022highresolutionimagesynthesislatent}. We additionally inject the clean datasets with \textbf{isotropic Gaussian noise} using $\sigma_\text{pix}=0.05$ on $[0,1]$-scaled images and apply a mild multiplicative color transform $(R,G,B)\mapsto(R,0.85G,0.70B)$ followed by brightness scaling of $0.85$ to simulate \textbf{camera browning}. Ground-truth labels remain unchanged so metrics are comparable to clean evaluation. Learned perception modules are trained on the corresponding transformed training images, so the evaluation measures covariance robustness with condition-matched perception rather than blind transfer to real weather. These adverse conditions are depicted in Figure~\ref{fig:weather_yolo}.

\section{Experiments}
\label{sec:experiments}

\subsection{Datasets}


\textbf{WildTrack~\cite{wildtrack}} contains 400 frames of 7 synchronized calibrated HD cameras (1920$\times$1080, 2\,fps) with ${\sim}$20 pedestrians per frame, covering a 12$\times$36\,m outdoor area.
\textbf{MultiviewX~\cite{multiviewx}} contains 400 frames of 6 synthetic cameras (1920$\times$1080, 2\,fps) with ${\sim}$40 pedestrians per frame, covering a 16$\times$25\,m area.


We use a contiguous 70/20/10 split on each dataset: frames 0--279 for CalibTrack training, 280--359 for validation, and 360--399 for testing. Within each dataset, condition, and tracker core, all covariance methods receive identical detections, confidences, depths, semantic embeddings, BEV measurements, and ground truth. Only the measurement covariance changes.  We fix $\alpha_{\mathrm{trk}}=\alpha_{\mathrm{xv}}=0.01$ for measurement-to-track association gates and cross-view fusion across all datasets, covariance methods, and tracker cores. Supp. Section~\ref{sec:gate_sensitivity} describes the gate parameterization and evaluates sensitivity to these settings.

On a single NVIDIA GeForce RTX 3090, the CalibTrack MLP forward pass averaged 0.002\,ms per detection (0.249\,ms and 0.217\,ms per frame WildTrack and MultiviewX respectively, highlighting minimal additional overhead for the tracking pipeline.

\subsection{Evaluation Metrics}
We report GOSPA~\cite{rahmathullah2017gospa} ($p{=}2$, $c{=}1$, $\alpha{=}2$), which penalizes false detections, missed targets, and localization error.
Additionally, we report MOT metrics~\cite{bernardin2008clear} and identity-aware IDF1~\cite{ristani2016idf1} with a 1.0\,m positive assignment threshold for the camera datasets.

For calibration we report the mean quadratic score $s=\ve^\top \hat R^{-1}\ve$, empirical $2\sigma$ coverage, and normalized measurement NEES~\cite{barshalom2001estimation}.
Bold and underlined entries indicate the \textbf{best} and \underline{second-best} values in all results tables. For GOSPA lower is better, and for calibration metrics \textit{best} means closest to the stated ideal.







\begin{table*}[!t]
{\centering
\caption{\textbf{Tracking performance on WildTrack and MultiviewX.} Bold and underlined entries indicate the \textbf{best} and \underline{second-best} mean value of a covariance method for each dataset and tracker core. Avg. Adverse averages the six non-clean conditions. CalibTrack reports mean $\pm$ standard deviation over ten training initialization seeds.}
\label{tab:main}
{\tiny
\setlength{\tabcolsep}{2pt}
\resizebox{\textwidth}{!}{%
\begin{tabular}{@{}lccc|ccc|ccc|ccc@{}}
\toprule
\multicolumn{1}{c}{} & \multicolumn{6}{c|}{\textbf{WildTrack}} & \multicolumn{6}{c}{\textbf{MultiviewX}} \\
\cmidrule(lr){2-7}\cmidrule(lr){8-13}
& \multicolumn{3}{c|}{\textbf{Clean}} & \multicolumn{3}{c|}{\textbf{Avg. Adverse}} & \multicolumn{3}{c|}{\textbf{Clean}} & \multicolumn{3}{c}{\textbf{Avg. Adverse}} \\
\cmidrule(lr){2-4}\cmidrule(lr){5-7}\cmidrule(lr){8-10}\cmidrule(lr){11-13}
Variant & IDF1 $\uparrow$ & MOTA $\uparrow$ & GOSPA $\downarrow$ & IDF1 $\uparrow$ & MOTA $\uparrow$ & GOSPA $\downarrow$ & IDF1 $\uparrow$ & MOTA $\uparrow$ & GOSPA $\downarrow$ & IDF1 $\uparrow$ & MOTA $\uparrow$ & GOSPA $\downarrow$ \\
\midrule
GM-PHD + Original ModTrack & \textbf{95.50} & \textbf{91.39} & \textbf{1.200} & \second{75.81} & \second{74.26} & \second{1.742} & \textbf{86.15} & \textbf{86.35} & \second{1.825} & \second{56.77} & \textbf{59.61} & 2.830 \\
GM-PHD + Confidence-scaled & 84.38 & 86.55 & 1.331 & 75.44 & 72.34 & 1.813 & 82.61 & 84.20 & 1.947 & 56.49 & 58.71 & 2.846 \\
GM-PHD + Pooled $R\times T$ & \second{91.16} & \second{89.29} & \second{1.216} & 73.20 & 71.60 & 1.800 & 82.29 & 84.67 & \textbf{1.796} & \textbf{57.40} & 58.60 & \second{2.808} \\
GM-PHD + \textbf{CalibTrack} & \pmcell{89.62}{2.06} & \pmcell{88.49}{0.98} & \pmcell{1.240}{0.055} & \textbf{\pmcell{77.59}{0.66}} & \textbf{\pmcell{75.02}{0.38}} & \textbf{\pmcell{1.730}{0.018}} & \second{\pmcell{83.90}{1.30}} & \second{\pmcell{84.85}{0.60}} & \pmcell{1.846}{0.060} & \pmcell{56.43}{0.53} & \second{\pmcell{59.43}{0.34}} & \textbf{\pmcell{2.800}{0.018}} \\
\midrule
pruned $\delta$-GLMB + Original ModTrack & \textbf{86.51} & \textbf{80.99} & \textbf{1.596} & \second{65.35} & \textbf{57.60} & \textbf{2.233} & 83.62 & \second{85.01} & 1.963 & 64.80 & \second{63.48} & 2.803 \\
pruned $\delta$-GLMB + Confidence-scaled & 77.67 & 71.43 & 1.848 & 61.99 & 43.66 & 2.585 & 80.64 & 76.24 & 2.349 & 64.46 & 60.84 & 2.873 \\
pruned $\delta$-GLMB + Pooled $R\times T$ & 76.47 & 70.69 & 1.852 & 60.38 & 39.72 & 2.666 & \textbf{84.45} & 83.87 & \second{1.959} & \second{65.35} & 62.78 & \textbf{2.788} \\
pruned $\delta$-GLMB + \textbf{CalibTrack} & \second{\pmcell{81.38}{2.39}} & \second{\pmcell{75.67}{1.70}} & \second{\pmcell{1.748}{0.061}} & \textbf{\pmcell{67.58}{0.65}} & \second{\pmcell{51.49}{0.63}} & \second{\pmcell{2.446}{0.018}} & \second{\pmcell{83.79}{1.55}} & \textbf{\pmcell{85.42}{1.30}} & \textbf{\pmcell{1.935}{0.071}} & \textbf{\pmcell{66.08}{0.35}} & \textbf{\pmcell{63.60}{0.48}} & \second{\pmcell{2.797}{0.026}} \\
\midrule
JIPDA + Original ModTrack & \textbf{90.08} & \textbf{86.97} & \textbf{1.505} & 72.09 & \second{65.69} & \textbf{2.025} & 85.74 & 85.07 & 2.007 & 63.49 & \textbf{64.76} & \textbf{2.745} \\
JIPDA + Confidence-scaled & 85.77 & 82.25 & 1.595 & \second{72.65} & 63.85 & 2.053 & \textbf{87.43} & \textbf{87.88} & \textbf{1.900} & \second{64.66} & 64.13 & 2.781 \\
JIPDA + Pooled $R\times T$ & 85.84 & 85.40 & 1.554 & 69.17 & 64.72 & 2.042 & \second{86.79} & 86.41 & 1.919 & \textbf{65.21} & \second{64.71} & \second{2.746} \\
JIPDA + \textbf{CalibTrack} & \second{\pmcell{89.20}{1.02}} & \second{\pmcell{85.79}{0.43}} & \second{\pmcell{1.546}{0.055}} & \textbf{\pmcell{75.51}{0.99}} & \textbf{\pmcell{66.83}{0.47}} & \second{\pmcell{2.036}{0.008}} & \pmcell{84.96}{0.97} & \second{\pmcell{86.82}{0.46}} & \second{\pmcell{1.919}{0.026}} & \pmcell{64.14}{0.51} & \pmcell{64.38}{0.21} & \pmcell{2.746}{0.014} \\
\bottomrule
\end{tabular}
}
}
}
\end{table*}

\subsection{Tracking Performance}
\label{sec:clean}
\begin{table*}[!htbp]
{\centering
\caption{\textbf{Paired Five-fold temporal cross-validation on WildTrack and MultiviewX.} Entries are observed mean paired effects $\pm$ standard deviation across five fold means. Each fold averages five training seeds and three tracker cores. Avg. Adverse additionally averages across the six conditions equally. Positive mean effects are shown in green and negative in red.}
\label{tab:aggregate_temporal_cv}
{\scriptsize
\setlength{\tabcolsep}{2pt}
\resizebox{\textwidth}{!}{%
\begin{tabular}{@{}lccc|ccc|ccc|ccc@{}}
\toprule
& \multicolumn{6}{c|}{\textbf{WildTrack}} & \multicolumn{6}{c}{\textbf{MultiviewX}} \\
\cmidrule(lr){2-7}\cmidrule(lr){8-13}
& \multicolumn{3}{c|}{\textbf{Clean}} & \multicolumn{3}{c|}{\textbf{Avg. Adverse}} & \multicolumn{3}{c|}{\textbf{Clean}} & \multicolumn{3}{c}{\textbf{Avg. Adverse}} \\
\cmidrule(lr){2-4}\cmidrule(lr){5-7}\cmidrule(lr){8-10}\cmidrule(lr){11-13}
Comparator
& $\Delta$IDF1 $\uparrow$ & $\Delta$MOTA $\uparrow$ & GOSPA benefit $\uparrow$
& $\Delta$IDF1 $\uparrow$ & $\Delta$MOTA $\uparrow$ & GOSPA benefit $\uparrow$
& $\Delta$IDF1 $\uparrow$ & $\Delta$MOTA $\uparrow$ & GOSPA benefit $\uparrow$
& $\Delta$IDF1 $\uparrow$ & $\Delta$MOTA $\uparrow$ & GOSPA benefit $\uparrow$ \\
\midrule
Confidence-scaled
& \poscell{2.81}{1.71} & \poscell{3.65}{2.51} & \poscell{0.154}{0.075}
& \poscell{3.59}{0.78} & \poscell{5.25}{1.92} & \poscell{0.134}{0.034}
& \poscell{0.50}{1.38} & \poscell{0.84}{0.78} & \poscell{0.036}{0.048}
& \poscell{1.95}{0.86} & \poscell{2.52}{0.91} & \poscell{0.091}{0.033} \\
Original ModTrack
& \negcell{-0.09}{0.80} & \negcell{-0.36}{1.27} & \poscell{0.008}{0.023}
& \poscell{2.62}{0.76} & \poscell{1.48}{1.29} & \poscell{0.030}{0.031}
& \poscell{0.71}{1.78} & \negcell{-0.13}{1.28} & \negcell{-0.002}{0.094}
& \poscell{1.63}{0.22} & \poscell{0.07}{0.19} & \poscell{0.010}{0.021} \\
Pooled $R\times T$
& \poscell{0.62}{1.41} & \poscell{1.13}{1.36} & \poscell{0.023}{0.029}
& \poscell{2.38}{1.77} & \poscell{2.94}{1.22} & \poscell{0.050}{0.023}
& \negcell{-0.97}{1.74} & \poscell{0.86}{1.11} & \negcell{-0.009}{0.082}
& \negcell{-0.81}{0.50} & \poscell{0.39}{0.31} & \negcell{-0.015}{0.017} \\
\bottomrule
\end{tabular}%
}
}
}
\end{table*}

We evaluate how CalibTrack’s learned covariance affects tracking performance under clean and adverse conditions across both datasets and three analytical tracker cores.

Table~\ref{tab:main} reports tracking results over the full 40-frame test window. We train ten CalibTrack heads per dataset with different random initializations and report the mean and standard deviation. The fixed covariance baselines are deterministic under the shared inputs and settings, so each has a single score. Average adverse results weight results equally across the six non-clean conditions.


Under clean conditions, Table~\ref{tab:main} shows that ModTrack’s hand-tuned covariance leads on WildTrack within each tracker variant and achieves the highest IDF1 and MOTA among covariance methods when paired with GM-PHD on MultiviewX.

ModTrack applies the same hand-tuned covariance rule across conditions, whereas CalibTrack learns heteroscedastic covariance that varies with each detection’s context, reducing reliance on manually designing and tuning this rule.  These results show that a carefully tuned covariance heuristic can remain stronger under clean conditions. However, CalibTrack’s primary focus is learning detection-dependent uncertainty to improve robustness under adverse conditions.


Under adverse conditions, Table~\ref{tab:main} shows that CalibTrack achieves the highest IDF1 among covariance methods within each of the three tracker cores on WildTrack, along with the best MOTA and GOSPA with GM-PHD. On MultiviewX, CalibTrack achieves the best IDF1 and MOTA among covariance methods with pruned $\delta$-GLMB and the best GOSPA with GM-PHD, while JIPDA results remain mixed.

These findings show that learned, detection-dependent covariance provides practical robustness benefits beyond confidence scaling and global recalibration on WildTrack, with additional advantages over hand-tuned covariance in specific metrics and conditions. On MultiviewX, the higher pedestrian density increases occlusions and association ambiguity, limiting the gains achievable through covariance estimation alone. Covariance can better represent the uncertainty of available detections, but cannot recover missed detections or fully resolve identity ambiguity between nearby pedestrians. This reduces the performance gains over comparative covariance methods seen on WildTrack.

Table~\ref{tab:aggregate_temporal_cv} reports results from five-fold temporal cross-validation. Each 400-frame dataset is divided into five consecutive 80-frame folds, with each fold used once as the test set to assess sensitivity to the choice of test window. Within each fold, CalibTrack is paired (See Supp. Section~\ref{sec:paired_protocol_supp}) with each baseline covariance method under the same evaluation conditions to measure its relative effect. CalibTrack improves average adverse condition tracking over confidence scaling on both datasets and over ModTrack and pooled scalar recalibration on WildTrack, supporting robustness to the test interval. Further details are provided in Supp. Section~\ref{sec:temporal}.

\begin{table*}[!th]
{\centering
\caption{\textbf{Leave-one-condition-out covariance transfer on WildTrack and MultiviewX.} Entries report paired effects of CalibTrack relative to each comparator as observed means $\pm$ standard deviations across five seed-wise averages. Each seed-wise average weights the six omitted conditions and three tracker cores equally. Positive mean effects are shown in dark green and negative mean effects in dark red.}
\label{tab:aggregate_loco}
{\scriptsize
\setlength{\tabcolsep}{2pt}
\begin{adjustbox}{max width=\textwidth, center=\textwidth}
\begin{tabular}{@{}lccc|ccc@{}}
\toprule
& \multicolumn{3}{c|}{\textbf{WildTrack}} & \multicolumn{3}{c}{\textbf{MultiviewX}} \\
\cmidrule(lr){2-4}\cmidrule(lr){5-7}
& \multicolumn{3}{c|}{\textbf{Avg. Adverse}} & \multicolumn{3}{c}{\textbf{Avg. Adverse}} \\
\cmidrule(lr){2-4}\cmidrule(lr){5-7}
Comparator
& $\Delta$IDF1 $\uparrow$ & $\Delta$MOTA $\uparrow$ & GOSPA benefit $\uparrow$
& $\Delta$IDF1 $\uparrow$ & $\Delta$MOTA $\uparrow$ & GOSPA benefit $\uparrow$ \\
\midrule
Confidence-scaled
& \poscell{3.21}{0.83} & \poscell{4.28}{0.46} & \poscell{0.074}{0.023}
& \poscell{0.15}{0.22} & \poscell{1.25}{0.12} & \poscell{0.051}{0.007} \\
Original ModTrack
& \poscell{2.16}{0.83} & \negcell{-1.62}{0.46} & \negcell{-0.077}{0.023}
& \poscell{0.33}{0.22} & \negcell{-0.13}{0.12} & \poscell{0.011}{0.007} \\
Pooled $R\times T$
& \poscell{5.49}{0.83} & \poscell{5.86}{0.46} & \poscell{0.100}{0.023}
& \negcell{-0.39}{0.22} & \poscell{0.73}{0.12} & \poscell{0.005}{0.007} \\
\bottomrule
\end{tabular}
\end{adjustbox}
}
}
\end{table*}

Additionally, leave-one-condition-out experiments evaluate the transfer of CalibTrack to unseen adverse conditions. For each adverse condition, we train and validate five randomly initialized CalibTrack heads on the other five adverse conditions pooled with clean data. Each head is then tested on the adverse condition omitted from training. The detector, depth model, and semantic model remain matched to the test condition. Table~\ref{tab:aggregate_loco} reports Avg. Adverse effects over three tracker cores and six omitted conditions. Across the adverse conditions, CalibTrack improves mean tracking scores on WildTrack over pooled rescaling and confidence methods, though ModTrack's hand-tuned covariance remains better in MOTA and GOSPA. On MultiviewX, CalibTrack improves over confidence scaling and is competitive with the other baselines. The results support meaningful cross-condition transfer.

Per-condition scores and adaptive-filter comparisons appear in Supp. Tables~\ref{tab:full_wild} and~\ref{tab:full_mvx}.


\begin{figure*}[!htbp]
    \centering
    \includegraphics[width=\linewidth]{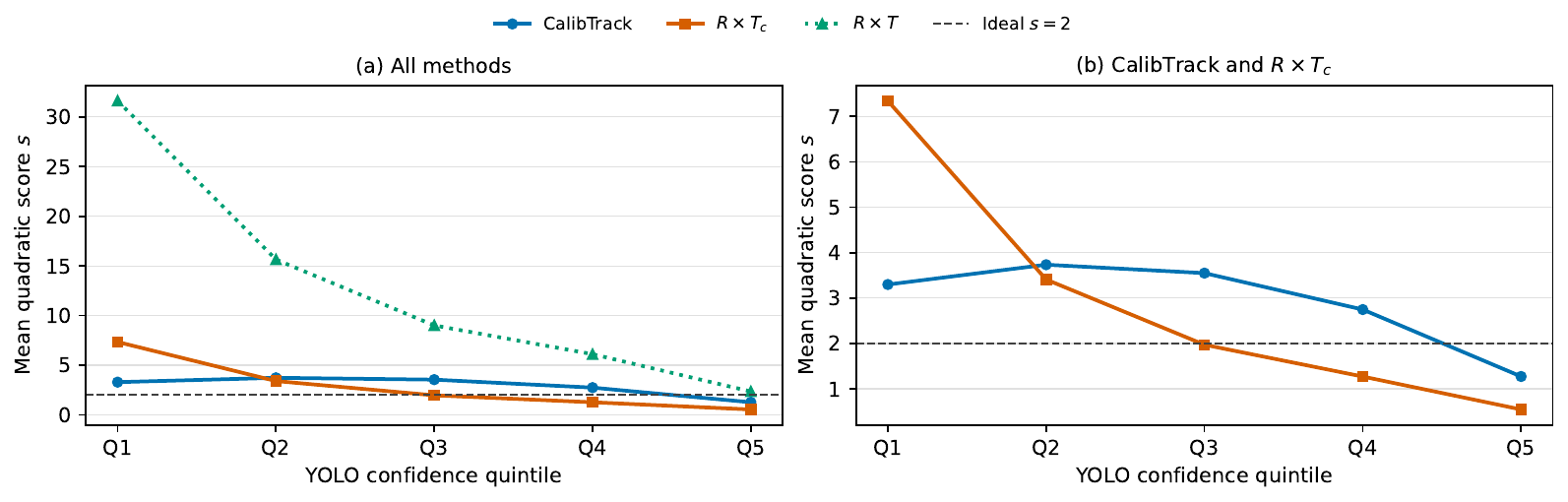}
    \caption{
    \textbf{Calibration across detection-confidence quintiles on WildTrack.} Mean quadratic scores are shown from the lowest (Q1) to highest (Q5) detection confidences. The ideal value is 2 for calibrated 2D Gaussian covariances. Higher values indicate overconfidence and lower values indicate underconfidence. Results averaged across adverse conditions and within confidence quintiles.}
    \label{fig:yolo_buckets}
\end{figure*}

\subsection{Measurement Calibration}

\begin{table}[!htbp]
\centering\scriptsize
\caption{\textbf{Measurement-level calibration.} Ideals: normalized measurement NEES $=1$ and $2\sigma$ coverage $=86.47\%$. Bold and underline mark the closest and second-closest point estimates to each ideal.}
\label{tab:calibration_summary}
\setlength{\tabcolsep}{4pt}
\begin{tabular}{@{}lcc|cc@{}}
\toprule
\multicolumn{5}{c}{\textbf{WildTrack}} \\
\midrule
 & \multicolumn{2}{c|}{\textbf{Clean}} & \multicolumn{2}{c}{\textbf{Avg. Adverse}} \\
\cmidrule(lr){2-3}\cmidrule(lr){4-5}
Covariance & NEES & $2\sigma$ Cov. (\%) & NEES & $2\sigma$ Cov. (\%) \\
\midrule
Original ModTrack & 0.451 & 94.42 & \textbf{0.863} & 89.76 \\
Confidence-scaled & \second{1.465} & \second{90.87} & 2.725 & \textbf{85.92} \\
Pooled $R\times T$ & 4.003 & 80.34 & 7.667 & 66.17 \\
\textbf{CalibTrack} & \textbf{1.127} & \textbf{90.02} & \second{1.582} & \second{85.04} \\
\midrule
\multicolumn{5}{c}{\textbf{MultiviewX}} \\
\midrule
 & \multicolumn{2}{c|}{\textbf{Clean}} & \multicolumn{2}{c}{\textbf{Avg. Adverse}} \\
\cmidrule(lr){2-3}\cmidrule(lr){4-5}
Covariance & NEES & $2\sigma$ Cov. (\%) & NEES & $2\sigma$ Cov. (\%) \\
\midrule
Original ModTrack & 0.370 & 95.55 & 0.593 & 92.61 \\
Confidence-scaled & \textbf{0.911} & 92.53 & \second{1.343} & \second{89.20} \\
Pooled $R\times T$ & 2.026 & \textbf{83.79} & 3.242 & 73.65 \\
\textbf{CalibTrack} & \second{0.804} & \second{91.47} & \textbf{1.033} & \textbf{87.77} \\
\bottomrule
\end{tabular}
\end{table}

We evaluate measurement covariance calibration using two complementary metrics: normalized measurement NEES and empirical $2\sigma$ coverage, computed from held-out detection residuals. Normalized NEES has an ideal value of $1$, because the expected squared Mahalanobis distance equals the measurement dimension when covariance is correctly calibrated. Empirical $2\sigma$ coverage has an ideal value of $86.47\%$, because a two-dimensional Gaussian places this fraction of its probability within the predicted two-sigma ellipse. Together, these metrics assess whether predicted uncertainty matches observed localization errors in average magnitude and probability.

Table~\ref{tab:calibration_summary} reports these calibration metrics on the test set of each dataset and shows CalibTrack is closest to both ideals on clean WildTrack. Under adverse conditions, Original ModTrack gives the closest WildTrack NEES and confidence scaling gives the closest coverage, which, however, coincides with a highly inflated NEES score indicating overconfidence. On MultiviewX, CalibTrack is closest to both ideals under adverse conditions, with NEES of $1.033$ and coverage of $87.77\%$. On clean frames, confidence scaling achieves the closest NEES, while pooled $R\times T$ achieves the closest coverage. However, their respective coverage of $92.53\%$ and NEES of $2.026$ reveal limitations that are obscured when either metric is considered alone. CalibTrack ranks second on both metrics for clean MultiviewX offering a balanced compromise between average error calibration and probability coverage. Full per-condition results are reported and discussed in Supp. Section~\ref{sec:measurement_calibration_supp}.

The WildTrack loss ablation in Supplementary Table~\ref{tab:loss_composition_conditions} shows that closer aggregate calibration does not necessarily correspond to better tracking. NLL alone gives average-adverse NEES and coverage closer to their targets, while the full objective gives better average-adverse tracking scores. This indicates that aggregate calibration and tracking performance can diverge, motivating their joint evaluation along with consideration of detection-level covariance when assessing measurement covariance estimates for downstream tracking performance.


\subsection{Robustness to Camera Removal}

\label{sec:camera_dropout}

We additionally evaluate CalibTrack with the identity-informed GM-PHD core under varying levels of camera dropout on the clean test splits of both datasets (Supp. Table~\ref{tab:camera_dropout}). Scores average all camera-removal configurations and ten trained heads. Because the learned module predicts covariance for each available detection, it accommodates changes in camera count without modification or retraining. Removing one camera reduces IDF1 from 89.62 to 88.31 on WildTrack and from 83.90 to 78.35 on MultiviewX. Larger reductions in coverage progressively impair tracking, with IDF1 reaching 31.40 and 34.62 when only two cameras remain. These results demonstrate that the same learned covariance module remains usable under camera dropout, while tracking performance depends on the number and geometry of the remaining views. Full details on the clean test set along with additional camera dropout results under adverse conditions are detailed in Supp. Section~\ref{sec:camera_dropout_supp}.

\subsection{Necessity of Heteroscedasticity}
To examine whether detection-dependent covariance improves calibration beyond condition-level rescaling we compare CalibTrack to the pooled $R\times T$ baseline, which applies the same scaled covariance to each detection. The latter's poor adverse condition calibration (Table~\ref{tab:calibration_summary}) could reflect differences in average covariance scale between conditions or its inability to adapt covariance to individual detections. To separate these effects, we introduce $R\times T_c$, fitting a distinct scalar $T_c$ per adverse condition. We group WildTrack detections from the test set within each adverse condition into five quintiles of increasing YOLO confidence, average the mean quadratic scores $s$ across the adverse conditions and compare the three methods in Figure~\ref{fig:yolo_buckets}. Values above the calibrated target of 2 indicate overconfidence and vice versa.

Pooled $R\times T$ produces heavily overconfident covariances for the lower confidence detections at 31 for the lowest confidence quintile. Condition specific $R\times T_c$ substantially reduces this overconfidence indicating the advantage of condition-level rescaling. However, $R\times T_c$ remains overconfident for the lowest confidence YOLO detections while also underconfident for the highest confidence detections. Meanwhile, CalibTrack's quadratic scores are less reactive to the confidence of the YOLO detections, producing better calibrated covariances on average. These findings support the benefit of CalibTrack's heteroscedastic predictions under adverse conditions. Supp. Section~\ref{sec:hetero} contains further analysis into per condition results.

\section{Conclusion}
\label{sec:conclusion}

This paper explores measurement covariance estimation methods and examines how covariance calibration affects analytical multi-camera multi-object tracking under clean and adverse conditions on WildTrack and MultiviewX. As a baseline, we study heuristic, confidence-scaled, and global scalar-recalibrated covariance approaches and introduce CalibTrack, a learned heteroscedastic module that predicts  per-detection covariance, under a training objective that encourages calibration. The gains over global recalibration on WildTrack show the benefit of adapting uncertainty to individual measurements, while the gains over confidence scaling suggest that detector confidence alone is insufficient for localization uncertainty. However, ModTrack's hand-tuned covariance remains the strongest baseline under clean conditions, suggesting that adaptive or learned covariance methods are unnecessary where the conditions are consistent and can be manually accounted for. The smaller and less consistent gains on MultiviewX point to a limit of covariance estimation. Specifically, it can help a tracker weigh the detections it receives, but its usefulness diminishes when occlusion removes detections or leaves associations hard to distinguish. Analyses across ten random initializations, five-fold temporal cross-validation, and leave-one-condition-out training support the robustness of our findings. Together, these analyses highlight the value of heteroscedastic covariance within analytical tracking, while showing that its benefits remain bounded by the quality of available observations and the downstream tracker. Future directions for this line of work include curating a multi-camera dataset captured under real adverse weather conditions, which is currently absent in the field. Code and trained point models will be made public upon acceptance.

\section{LLM Usage Statement}

    We used Codex to assist with code generation and experiment execution, and ChatGPT to provide suggestions on table formatting and LaTeX layout. All manuscript text was written by the authors, with AI assistance limited to suggestions for wording. The authors manually reviewed the generated code and take full responsibility for the final manuscript, code, and reported results. LLMs were not used to develop research ideas, design experiments, or to review related work. All conceptual contributions and experimental designs originated with the authors.

\FloatBarrier
{
    \small
    \bibliographystyle{unsrt}
    \bibliography{references}
}

\clearpage
\appendix

\section*{Supplementary Material}
\addcontentsline{toc}{section}{Supplementary Material}

\section{Covariance Head: Inputs and Camera Identity Ablation}
\label{sec:covariance_head_supp}

\subsection{Input Features}

Each per-camera detection produces a 10-dimensional camera-context vector and the covariance head maps this vector to three Cholesky parameters for a positive-definite $2\times2$ covariance matrix.

The 10 camera features are as follows and are detailed in ModTrack~\cite{iyer2025modtrack}:
\begin{itemize}
    \item \textbf{Depth}: absolute projection depth used to project the detection footpoint to BEV. Farther detections usually have larger localization uncertainty.
    \item \textbf{Box Height}: detector-box height in pixels, used as a scale cue and proxy for range.
    \item \textbf{Box Width}: detector-box width in pixels, which helps identify unusual boxes, partial visibility, or merged detections.
    \item \textbf{Detection Confidence}: YOLO confidence clipped to $[0,1]$, used as a detector reliability cue.
    \item \textbf{Geometric Depth Estimate Disagreement}: absolute disagreement between footpoint-ray depth and the bounding-box height depth proxy.
    \item \textbf{Lift Depth}: Lift-estimated depth aggregated over sampled detection-box points.
    \item \textbf{Lift Depth Disagreement}: absolute disagreement between Lift depth and the precision weighted depth (from the footpoint-ray projection and the bounding-box height cues).
    \item \textbf{Normalized Footpoint u}: horizontal projected footpoint coordinate divided by image width.
    \item \textbf{Normalized Footpoint v}: vertical projected footpoint coordinate divided by image height.
    \item \textbf{Camera ID}: camera ID normalized by the number of cameras minus one, allowing camera-specific calibration biases.
\end{itemize}

\subsection{Implementation and Training Details}

\textbf{Architecture and preprocessing.}
The covariance head is implemented in PyTorch 1.12~\cite{pytorch} as three fully connected layers with dimensions $10\!\to\!64\!\to\!64\!\to\!3$, with ReLU activations after the two hidden layers and a linear output layer. Depth features use absolute depths and depth-cue disagreements in metres. Box height and width remain in pixels and are lower-bounded by one pixel. Detector confidence and normalized footpoint coordinates are clipped to $[0,1]$, and camera identity is divided by the number of cameras minus one. Unavailable auxiliary depth estimates are replaced by the projection depth. These deterministically preprocessed features are passed directly to the MLP without training-set mean or variance normalization.

\textbf{Supervision and optimization.}
We train one head per dataset using pooled clean, rain, snow, fog, dust, Gaussian-noise, and browning supervision. Each projected detection is matched independently to its nearest ground-truth BEV point within 3\,m, without enforcing one-to-one assignment or using covariance-dependent gates. The primary split uses frames 0--279 for training, 280--359 for validation, and 360--399 for testing. Perception outputs and the shared pose covariance remain fixed during covariance-head training. Condition labels are used to group the loss and are not inputs to the head.

\subsection{Camera Identity Ablation}
\label{sec:camera_id_ablation}

Camera identity is one of the ten context features supplied to the covariance head. To assess the contribution of camera identity on the fixed cameras, we remove it, giving a nine-feature input to the head. We retrain both variants with 10 seeds (42--51) using identical supervision, loss, and training budget. Perception caches, the contiguous 70/20/10 split, test frames 360--399, and gates $(0.01,0.01)$ remain fixed. Each reported score weights the six adverse conditions equally.

\begin{table*}[!tbp]
\centering
\scriptsize
\caption{\textbf{Camera-ID ablation.} Full and No ID entries are mean
$\pm$ standard deviation across ten training seeds. The six adverse
conditions are averaged within each seed. Differences are mean
no ID minus full for matched seeds.
Positive differences are green. Negative differences are red.}
\label{tab:camera_id_ablation}
\setlength{\tabcolsep}{3pt}
\renewcommand{\arraystretch}{1.08}

\begin{minipage}[t]{0.49\textwidth}
\centering
\textbf{(a) WildTrack}\par
\vspace{4pt}
\begin{adjustbox}{max width=\linewidth}
\begin{tabular}{lrrr}
\toprule
Metric & Full (10) & No ID (9) & Difference \\
\midrule
\multicolumn{4}{l}{\textbf{GM-PHD}} \\
\addlinespace[2pt]
IDF1
 & \pmcell{77.75}{1.58}
 & \pmcell{77.61}{0.51}
 & \negdiff{-0.135} \\
MOTA
 & \pmcell{74.86}{1.05}
 & \pmcell{74.85}{0.51}
 & \negdiff{-0.016} \\
GOSPA
 & \pmcell{1.751}{0.044}
 & \pmcell{1.725}{0.017}
 & \negdiff{-0.0254} \\
\midrule
\multicolumn{4}{l}{\textbf{pruned $\delta$-GLMB}} \\
\addlinespace[2pt]
IDF1
 & \pmcell{67.67}{0.74}
 & \pmcell{68.22}{0.74}
 & \posdiff{0.546} \\
MOTA
 & \pmcell{51.48}{0.75}
 & \pmcell{51.75}{0.54}
 & \posdiff{0.273} \\
GOSPA
 & \pmcell{2.455}{0.030}
 & \pmcell{2.440}{0.018}
 & \negdiff{-0.0154} \\
\midrule
\multicolumn{4}{l}{\textbf{JIPDA}} \\
\addlinespace[2pt]
IDF1
 & \pmcell{75.04}{0.97}
 & \pmcell{75.78}{0.81}
 & \posdiff{0.740} \\
MOTA
 & \pmcell{66.74}{0.56}
 & \pmcell{66.81}{0.39}
 & \posdiff{0.070} \\
GOSPA
 & \pmcell{2.051}{0.024}
 & \pmcell{2.033}{0.010}
 & \negdiff{-0.0178} \\
\bottomrule
\end{tabular}
\end{adjustbox}
\end{minipage}
\hfill
\begin{minipage}[t]{0.49\textwidth}
\centering
\textbf{(b) MultiviewX}\par
\vspace{4pt}
\begin{adjustbox}{max width=\linewidth}
\begin{tabular}{lrrr}
\toprule
Metric & Full (10) & No ID (9) & Difference \\
\midrule
\multicolumn{4}{l}{\textbf{GM-PHD}} \\
\addlinespace[2pt]
IDF1
 & \pmcell{56.25}{0.51}
 & \pmcell{56.86}{0.60}
 & \posdiff{0.609} \\
MOTA
 & \pmcell{59.43}{0.30}
 & \pmcell{59.48}{0.34}
 & \posdiff{0.054} \\
GOSPA
 & \pmcell{2.814}{0.027}
 & \pmcell{2.800}{0.012}
 & \negdiff{-0.0137} \\
\midrule
\multicolumn{4}{l}{\textbf{pruned $\delta$-GLMB}} \\
\addlinespace[2pt]
IDF1
 & \pmcell{65.43}{0.39}
 & \pmcell{65.95}{0.64}
 & \posdiff{0.521} \\
MOTA
 & \pmcell{62.86}{0.96}
 & \pmcell{63.12}{0.86}
 & \posdiff{0.260} \\
GOSPA
 & \pmcell{2.827}{0.041}
 & \pmcell{2.809}{0.031}
 & \negdiff{-0.0178} \\
\midrule
\multicolumn{4}{l}{\textbf{JIPDA}} \\
\addlinespace[2pt]
IDF1
 & \pmcell{64.16}{0.42}
 & \pmcell{64.28}{0.46}
 & \posdiff{0.120} \\
MOTA
 & \pmcell{64.24}{0.29}
 & \pmcell{64.49}{0.25}
 & \posdiff{0.244} \\
GOSPA
 & \pmcell{2.759}{0.025}
 & \pmcell{2.747}{0.012}
 & \negdiff{-0.0125} \\
\bottomrule
\end{tabular}
\end{adjustbox}
\end{minipage}
\end{table*}


Table~\ref{tab:camera_id_ablation} reports observed mean scores and paired differences, with standard deviations across ten training seeds. Removing camera identity changes average IDF1 by at most 0.74, MOTA by at most 0.273, and GOSPA by at most 0.0254.

\section{Covariance Loss Composition Ablation}

To justify the choice of loss (NLL + NEES + coverage) we ablate the calibrated covariance objective on WildTrack using the ModTrack identity-informed GM-PHD tracker core across all conditions. Table~\ref{tab:loss_composition_conditions} reports the per-condition mean and standard deviation across three seeds for this ablation. All runs use the same mixed WildTrack covariance supervision, object detector, depth model, feature extractor, and tracker hyperparameters, including $\alpha_{\mathrm{trk}}=\alpha_{\mathrm{xv}}=0.01$. Only the covariance-head loss is varied. Bold and underline mark the best and second-best entry for each condition group. Calibration columns are ranked by closeness to their ideals (mean $s=2.0$, 2$\sigma$ coverage $=86.5\%$, and normalized NEES $=1.0$).

NLL alone gives average-adverse NEES and $2\sigma$ coverage closest to their nominal targets. However, the full NLL + NEES + coverage loss achieves the best average-adverse results across all reported tracking metrics, although it does not lead in every condition. We therefore use the full three-component loss in the main paper based on its empirically superior tracking performance.

\begin{table*}[!thbp]
\centering
\caption{\textbf{Per-condition WildTrack loss-composition ablation.} Entries are mean $\pm$ standard deviation over three covariance-head training seeds. Rows are grouped by condition, and \textbf{bold}/\underline{underline} ranks are computed within each condition group. Avg. Adverse averages the six non-clean conditions per seed.}
\label{tab:loss_composition_conditions}
{\tiny
\setlength{\tabcolsep}{2pt}
\resizebox{\textwidth}{!}{%
\begin{tabular}{llrrrrrrrrr}\toprule
Condition & Loss & IDF1 & MOTA & HOTA & AssA & DetA & GOSPA & IDS & NEES & $2\sigma$ cov. \\
\midrule
Clean & NLL & \pmcell{89.46}{2.62} & \pmcell{87.96}{1.79} & \pmcell{71.17}{4.22} & \pmcell{68.59}{4.65} & \pmcell{74.02}{4.07} & \pmcell{1.351}{0.144} & \second{\pmcell{8.00}{4.58}} & \textbf{\pmcell{1.017}{0.048}} & \pmcell{91.03}{0.26} \\
Clean & NLL + NEES & \pmcell{89.19}{2.42} & \textbf{\pmcell{89.29}{0.82}} & \pmcell{73.24}{1.21} & \pmcell{68.97}{2.80} & \textbf{\pmcell{77.92}{0.70}} & \second{\pmcell{1.209}{0.012}} & \pmcell{8.67}{1.53} & \pmcell{1.096}{0.005} & \textbf{\pmcell{90.22}{0.13}} \\
Clean & NLL + Coverage & \textbf{\pmcell{91.42}{1.32}} & \second{\pmcell{89.25}{1.07}} & \textbf{\pmcell{74.95}{1.81}} & \textbf{\pmcell{74.71}{1.76}} & \pmcell{75.32}{3.19} & \pmcell{1.298}{0.114} & \textbf{\pmcell{7.33}{1.53}} & \second{\pmcell{1.068}{0.053}} & \pmcell{90.48}{0.43} \\
Clean & NLL + NEES + Coverage & \second{\pmcell{89.74}{1.27}} & \pmcell{88.97}{0.18} & \second{\pmcell{73.65}{1.01}} & \second{\pmcell{69.91}{1.83}} & \second{\pmcell{77.71}{0.09}} & \textbf{\pmcell{1.202}{0.009}} & \pmcell{11.33}{2.52} & \pmcell{1.100}{0.082} & \second{\pmcell{90.27}{0.55}} \\
\midrule
Rain & NLL & \second{\pmcell{68.75}{5.15}} & \pmcell{66.04}{4.36} & \pmcell{46.97}{4.04} & \second{\pmcell{39.42}{4.86}} & \pmcell{56.17}{2.73} & \pmcell{2.035}{0.150} & \pmcell{38.00}{7.21} & \textbf{\pmcell{2.195}{0.108}} & \textbf{\pmcell{82.01}{1.12}} \\
Rain & NLL + NEES & \pmcell{67.90}{3.04} & \pmcell{66.32}{0.76} & \second{\pmcell{47.56}{1.63}} & \pmcell{38.50}{3.34} & \second{\pmcell{59.07}{1.15}} & \second{\pmcell{1.964}{0.045}} & \pmcell{40.67}{0.58} & \second{\pmcell{2.226}{0.082}} & \pmcell{81.08}{1.02} \\
Rain & NLL + Coverage & \pmcell{67.43}{1.04} & \textbf{\pmcell{67.89}{0.79}} & \pmcell{47.47}{2.11} & \pmcell{39.36}{2.35} & \pmcell{57.39}{1.86} & \pmcell{1.978}{0.062} & \textbf{\pmcell{35.00}{4.00}} & \pmcell{2.258}{0.097} & \second{\pmcell{81.70}{0.16}} \\
Rain & NLL + NEES + Coverage & \textbf{\pmcell{70.17}{1.71}} & \second{\pmcell{67.37}{2.64}} & \textbf{\pmcell{50.13}{1.97}} & \textbf{\pmcell{42.41}{2.63}} & \textbf{\pmcell{59.43}{1.37}} & \textbf{\pmcell{1.955}{0.057}} & \second{\pmcell{35.33}{5.69}} & \pmcell{2.263}{0.158} & \pmcell{81.46}{0.65} \\
\midrule
Snow & NLL & \pmcell{77.03}{4.28} & \pmcell{75.04}{5.92} & \pmcell{55.78}{6.31} & \pmcell{49.49}{6.48} & \pmcell{63.04}{6.05} & \pmcell{1.788}{0.236} & \pmcell{28.33}{10.69} & \textbf{\pmcell{1.600}{0.077}} & \textbf{\pmcell{85.86}{0.56}} \\
Snow & NLL + NEES & \textbf{\pmcell{79.59}{1.69}} & \second{\pmcell{78.12}{0.80}} & \textbf{\pmcell{59.65}{0.57}} & \textbf{\pmcell{53.39}{0.77}} & \second{\pmcell{66.89}{0.40}} & \second{\pmcell{1.618}{0.008}} & \pmcell{21.67}{2.89} & \pmcell{1.672}{0.079} & \pmcell{84.48}{0.88} \\
Snow & NLL + Coverage & \second{\pmcell{79.44}{1.45}} & \textbf{\pmcell{78.40}{0.95}} & \pmcell{57.88}{3.73} & \second{\pmcell{51.19}{4.75}} & \pmcell{65.62}{2.47} & \pmcell{1.659}{0.071} & \second{\pmcell{20.33}{4.93}} & \second{\pmcell{1.641}{0.054}} & \second{\pmcell{85.49}{0.23}} \\
Snow & NLL + NEES + Coverage & \pmcell{79.20}{1.38} & \textbf{\pmcell{78.40}{0.16}} & \second{\pmcell{58.00}{1.00}} & \pmcell{49.76}{1.54} & \textbf{\pmcell{67.82}{0.21}} & \textbf{\pmcell{1.591}{0.004}} & \textbf{\pmcell{19.67}{2.52}} & \pmcell{1.691}{0.107} & \pmcell{84.69}{0.79} \\
\midrule
Fog & NLL & \pmcell{76.11}{3.26} & \pmcell{71.04}{3.12} & \pmcell{54.17}{5.31} & \second{\pmcell{48.98}{7.00}} & \pmcell{60.23}{3.34} & \pmcell{1.904}{0.131} & \pmcell{41.33}{5.51} & \textbf{\pmcell{1.406}{0.073}} & \textbf{\pmcell{85.73}{0.71}} \\
Fog & NLL + NEES & \pmcell{75.94}{3.73} & \pmcell{71.92}{1.61} & \pmcell{53.75}{5.99} & \pmcell{46.80}{9.92} & \second{\pmcell{62.47}{1.16}} & \second{\pmcell{1.822}{0.064}} & \second{\pmcell{36.00}{3.00}} & \pmcell{1.503}{0.065} & \pmcell{84.26}{0.72} \\
Fog & NLL + Coverage & \second{\pmcell{78.39}{2.72}} & \second{\pmcell{73.11}{3.29}} & \second{\pmcell{54.48}{1.43}} & \pmcell{48.80}{1.16} & \pmcell{61.09}{2.08} & \pmcell{1.864}{0.079} & \pmcell{36.33}{7.51} & \second{\pmcell{1.450}{0.040}} & \second{\pmcell{85.09}{0.19}} \\
Fog & NLL + NEES + Coverage & \textbf{\pmcell{78.46}{1.36}} & \textbf{\pmcell{74.75}{0.52}} & \textbf{\pmcell{58.39}{2.49}} & \textbf{\pmcell{54.47}{4.81}} & \textbf{\pmcell{63.08}{0.29}} & \textbf{\pmcell{1.783}{0.007}} & \textbf{\pmcell{27.67}{2.52}} & \pmcell{1.509}{0.084} & \pmcell{84.43}{0.76} \\
\midrule
Dust & NLL & \pmcell{72.49}{6.50} & \pmcell{69.75}{4.64} & \pmcell{51.98}{7.43} & \pmcell{45.38}{10.12} & \pmcell{60.11}{3.53} & \pmcell{1.898}{0.160} & \pmcell{37.00}{15.13} & \textbf{\pmcell{1.509}{0.079}} & \textbf{\pmcell{86.21}{0.76}} \\
Dust & NLL + NEES & \second{\pmcell{75.03}{2.18}} & \textbf{\pmcell{74.37}{1.10}} & \textbf{\pmcell{55.70}{0.90}} & \textbf{\pmcell{49.49}{1.17}} & \second{\pmcell{63.01}{1.28}} & \second{\pmcell{1.759}{0.032}} & \textbf{\pmcell{27.33}{4.93}} & \pmcell{1.616}{0.078} & \pmcell{84.53}{1.15} \\
Dust & NLL + Coverage & \pmcell{72.42}{3.15} & \pmcell{71.88}{2.17} & \pmcell{53.00}{1.65} & \pmcell{46.36}{2.13} & \pmcell{60.92}{1.25} & \pmcell{1.847}{0.066} & \pmcell{31.67}{4.51} & \second{\pmcell{1.554}{0.065}} & \second{\pmcell{85.58}{0.34}} \\
Dust & NLL + NEES + Coverage & \textbf{\pmcell{76.09}{0.75}} & \second{\pmcell{73.91}{0.42}} & \second{\pmcell{55.25}{2.38}} & \second{\pmcell{48.16}{4.36}} & \textbf{\pmcell{63.77}{0.23}} & \textbf{\pmcell{1.757}{0.017}} & \second{\pmcell{31.00}{3.46}} & \pmcell{1.624}{0.093} & \pmcell{84.96}{0.76} \\
\midrule
Gauss $\sigma{=}0.05$ & NLL & \pmcell{76.52}{5.19} & \pmcell{69.71}{2.54} & \pmcell{56.75}{5.19} & \pmcell{54.80}{6.86} & \pmcell{59.04}{3.39} & \pmcell{1.938}{0.143} & \pmcell{25.00}{6.93} & \textbf{\pmcell{1.129}{0.074}} & \second{\pmcell{87.22}{0.73}} \\
Gauss $\sigma{=}0.05$ & NLL + NEES & \pmcell{76.94}{1.53} & \pmcell{72.06}{0.73} & \textbf{\pmcell{60.14}{1.23}} & \textbf{\pmcell{58.62}{2.11}} & \second{\pmcell{61.95}{0.78}} & \second{\pmcell{1.817}{0.028}} & \pmcell{22.00}{1.00} & \pmcell{1.315}{0.102} & \pmcell{85.29}{0.54} \\
Gauss $\sigma{=}0.05$ & NLL + Coverage & \second{\pmcell{77.86}{1.39}} & \second{\pmcell{72.30}{1.21}} & \pmcell{58.32}{2.24} & \pmcell{56.26}{3.63} & \pmcell{60.70}{1.11} & \pmcell{1.847}{0.041} & \second{\pmcell{19.33}{1.53}} & \second{\pmcell{1.154}{0.009}} & \textbf{\pmcell{86.60}{0.18}} \\
Gauss $\sigma{=}0.05$ & NLL + NEES + Coverage & \textbf{\pmcell{78.19}{0.51}} & \textbf{\pmcell{73.14}{0.16}} & \second{\pmcell{59.45}{0.97}} & \second{\pmcell{56.95}{1.94}} & \textbf{\pmcell{62.28}{0.12}} & \textbf{\pmcell{1.801}{0.017}} & \textbf{\pmcell{19.00}{2.65}} & \pmcell{1.305}{0.065} & \pmcell{85.53}{0.67} \\
\midrule
Browning & NLL & \pmcell{86.03}{2.77} & \pmcell{83.19}{2.52} & \pmcell{66.83}{3.69} & \pmcell{64.44}{4.81} & \pmcell{69.46}{3.14} & \pmcell{1.539}{0.149} & \textbf{\pmcell{16.00}{4.00}} & \textbf{\pmcell{0.997}{0.068}} & \pmcell{90.87}{0.40} \\
Browning & NLL + NEES & \textbf{\pmcell{86.55}{2.09}} & \second{\pmcell{84.31}{0.32}} & \textbf{\pmcell{68.97}{1.94}} & \textbf{\pmcell{66.31}{3.45}} & \second{\pmcell{71.93}{0.35}} & \second{\pmcell{1.437}{0.005}} & \textbf{\pmcell{16.00}{2.00}} & \pmcell{1.040}{0.030} & \second{\pmcell{90.12}{0.32}} \\
Browning & NLL + Coverage & \second{\pmcell{86.21}{2.61}} & \textbf{\pmcell{84.38}{0.52}} & \pmcell{68.26}{1.85} & \second{\pmcell{65.83}{4.41}} & \pmcell{71.02}{1.33} & \pmcell{1.488}{0.048} & \textbf{\pmcell{16.00}{1.73}} & \second{\pmcell{1.018}{0.038}} & \pmcell{90.48}{0.11} \\
Browning & NLL + NEES + Coverage & \pmcell{84.62}{2.28} & \second{\pmcell{84.31}{0.34}} & \second{\pmcell{68.50}{1.51}} & \pmcell{64.83}{2.88} & \textbf{\pmcell{72.53}{0.30}} & \textbf{\pmcell{1.426}{0.009}} & \second{\pmcell{16.33}{0.58}} & \pmcell{1.068}{0.062} & \textbf{\pmcell{90.08}{0.31}} \\
\midrule
Avg. adverse & NLL & \pmcell{76.15}{4.14} & \pmcell{72.46}{3.75} & \pmcell{55.41}{5.09} & \pmcell{50.42}{6.22} & \pmcell{61.34}{3.60} & \pmcell{1.850}{0.160} & \pmcell{30.94}{6.85} & \textbf{\pmcell{1.472}{0.079}} & \textbf{\pmcell{86.32}{0.70}} \\
Avg. adverse & NLL + NEES & \second{\pmcell{76.99}{0.72}} & \pmcell{74.52}{0.27} & \second{\pmcell{57.63}{1.36}} & \second{\pmcell{52.19}{2.11}} & \second{\pmcell{64.22}{0.79}} & \second{\pmcell{1.736}{0.023}} & \pmcell{27.28}{0.51} & \pmcell{1.562}{0.071} & \pmcell{84.96}{0.76} \\
Avg. adverse & NLL + Coverage & \pmcell{76.96}{0.83} & \second{\pmcell{74.66}{0.59}} & \pmcell{56.57}{1.46} & \pmcell{51.30}{1.35} & \pmcell{62.79}{1.61} & \pmcell{1.781}{0.056} & \second{\pmcell{26.44}{1.75}} & \second{\pmcell{1.513}{0.048}} & \second{\pmcell{85.82}{0.13}} \\
Avg. adverse & NLL + NEES + Coverage & \textbf{\pmcell{77.79}{0.13}} & \textbf{\pmcell{75.32}{0.48}} & \textbf{\pmcell{58.29}{0.22}} & \textbf{\pmcell{52.76}{0.39}} & \textbf{\pmcell{64.82}{0.13}} & \textbf{\pmcell{1.719}{0.007}} & \textbf{\pmcell{24.83}{0.29}} & \pmcell{1.577}{0.093} & \pmcell{85.19}{0.64} \\
\bottomrule
\end{tabular}
}
}
\end{table*}

\section{Association Gates Parametrization and Sensitivity} 
\label{sec:gate_sensitivity}

CalibTrack's learned covariance influences downstream association, which is guided by a Mahalanobis gate. To ensure that the gate complements the calibrated covariance, we parameterize the association gate by a tail probability $\alpha$, representing the probability that a valid measurement lies outside the gate. For a two-dimensional measurement, the squared Mahalanobis distance follows a $\chi^2$ distribution with two degrees of freedom, giving
$\tau(\alpha) = -2\log \alpha,$
where $\tau$ is the gating threshold. The original ModTrack threshold $\tau=9.21$ corresponds to $\alpha=0.01$, i.e., a $99\%$ confidence gate.
We use the same tail probabilities $\alpha_{\mathrm{trk}}=\alpha_{\mathrm{xv}}=0.01$ for measurement-to-track association and cross-view fusion, across all datasets, every covariance method, and all tracker cores. These settings remain fixed for the main comparisons and supplementary tracking ablations. Supp. Section~\ref{sec:gate_sensitivity} quantifies sensitivity of both probabilities.

To quantify sensitivity to these parameters, we evaluate all 16 independent combinations in $\{0.005,0.01,0.025,0.05\}$. Table~\ref{tab:alpha_gate_ablation} reports the results for the default settings ($\alpha_{\mathrm{trk}}=\alpha_{\mathrm{xv}}=0.01$) along with the minimum--maximum range across the swept hyperparameters. The sweep covers both datasets, all three tracker cores, and all four covariance methods on clean data and the six adverse conditions. Perception caches, covariance checkpoints, backend settings, and test frames 360--399 remain fixed. Each pair uses the same ten CalibTrack heads and one evaluation per deterministic baseline.

Sensitivity sweeps in Table~\ref{tab:alpha_gate_ablation} show that the largest IDF1 ranges across the 16 gate pairs occur with pooled $R\times T$ covariance on WildTrack, reaching 5.34 points under clean conditions with GM-PHD and 3.23 points under average adverse conditions with JIPDA. For CalibTrack, the largest ranges are 2.12 and 1.68 points, respectively, both with GM-PHD on WildTrack. These results indicate that CalibTrack exhibits its greatest gate sensitivity with GM-PHD on WildTrack across all three metrics and both condition groups tested. Ultimately, this indicates low sensitivity to the hyperparameters being evaluated.

\begin{table*}[!thbp]
\centering
\caption{\textbf{Association-gate sensitivity on both datasets.} Each cell is the result using the default hyperparameters $(0.01,0.01)$ followed by [minimum, maximum] over all 16 gate pairs.}
\label{tab:alpha_gate_ablation}
{\scriptsize
\setlength{\tabcolsep}{2pt}
\resizebox{\textwidth}{!}{%
\begin{tabular}{lllccc|ccc}\toprule
Dataset & Core & Covariance & \multicolumn{3}{c|}{Clean} & \multicolumn{3}{c}{Avg. adverse} \\
 & & & IDF1 & MOTA & GOSPA & IDF1 & MOTA & GOSPA \\
WildTrack & GM-PHD & Original ModTrack & 95.50 [95.23, 95.50] & 91.39 [91.18, 91.39] & 1.200 [1.200, 1.206] & 75.81 [75.63, 76.73] & 74.26 [74.12, 75.18] & 1.742 [1.740, 1.746] \\
WildTrack & GM-PHD & Confidence-scaled & 84.38 [81.94, 84.59] & 86.55 [85.71, 86.55] & 1.331 [1.310, 1.331] & 75.44 [75.07, 75.44] & 72.34 [71.67, 72.51] & 1.813 [1.807, 1.824] \\
WildTrack & GM-PHD & Pooled $R\times T$ & 91.16 [87.21, 92.55] & 89.29 [87.50, 89.29] & 1.216 [1.213, 1.232] & 73.20 [72.48, 74.45] & 71.60 [69.73, 72.13] & 1.800 [1.783, 1.828] \\
WildTrack & GM-PHD & CalibTrack & 89.62 [89.62, 91.74] & 88.49 [88.38, 89.24] & 1.240 [1.226, 1.246] & 77.59 [77.59, 79.27] & 75.02 [74.48, 75.05] & 1.730 [1.727, 1.749] \\
WildTrack & pruned $\delta$-GLMB & Original ModTrack & 86.51 [86.51, 86.55] & 80.99 [80.99, 81.09] & 1.596 [1.593, 1.596] & 65.35 [65.35, 65.91] & 57.60 [57.60, 58.05] & 2.233 [2.229, 2.233] \\
WildTrack & pruned $\delta$-GLMB & Confidence-scaled & 77.67 [77.65, 77.67] & 71.43 [69.01, 71.43] & 1.848 [1.848, 1.907] & 61.99 [61.99, 62.06] & 43.66 [43.42, 43.66] & 2.585 [2.585, 2.591] \\
WildTrack & pruned $\delta$-GLMB & Pooled $R\times T$ & 76.47 [75.63, 76.50] & 70.69 [70.69, 70.90] & 1.852 [1.838, 1.852] & 60.38 [60.38, 60.64] & 39.72 [39.71, 40.06] & 2.666 [2.662, 2.666] \\
WildTrack & pruned $\delta$-GLMB & CalibTrack & 81.38 [81.38, 81.38] & 75.67 [75.67, 75.67] & 1.748 [1.748, 1.748] & 67.58 [67.58, 67.62] & 51.49 [51.49, 51.53] & 2.446 [2.446, 2.447] \\
WildTrack & JIPDA & Original ModTrack & 90.08 [90.08, 90.08] & 86.97 [86.97, 86.97] & 1.505 [1.505, 1.505] & 72.09 [72.04, 72.09] & 65.69 [65.69, 65.93] & 2.025 [2.025, 2.027] \\
WildTrack & JIPDA & Confidence-scaled & 85.77 [85.77, 85.99] & 82.25 [82.25, 83.40] & 1.595 [1.595, 1.614] & 72.65 [72.07, 72.65] & 63.85 [63.85, 64.13] & 2.053 [2.046, 2.053] \\
WildTrack & JIPDA & Pooled $R\times T$ & 85.84 [85.01, 86.22] & 85.40 [84.87, 85.82] & 1.554 [1.515, 1.557] & 69.17 [69.17, 72.40] & 64.72 [64.23, 65.74] & 2.042 [2.016, 2.042] \\
WildTrack & JIPDA & CalibTrack & 89.20 [89.20, 89.53] & 85.79 [85.79, 86.07] & 1.546 [1.546, 1.548] & 75.51 [75.51, 75.53] & 66.83 [66.73, 66.83] & 2.036 [2.036, 2.037] \\
MultiviewX & GM-PHD & Original ModTrack & 86.15 [85.93, 86.15] & 86.35 [85.54, 86.35] & 1.825 [1.825, 1.842] & 56.77 [56.65, 57.00] & 59.61 [59.50, 60.01] & 2.830 [2.814, 2.832] \\
MultiviewX & GM-PHD & Confidence-scaled & 82.61 [80.14, 82.61] & 84.20 [83.67, 84.20] & 1.947 [1.921, 1.960] & 56.49 [56.49, 58.30] & 58.71 [58.36, 58.75] & 2.846 [2.846, 2.856] \\
MultiviewX & GM-PHD & Pooled $R\times T$ & 82.29 [81.43, 83.89] & 84.67 [84.14, 84.67] & 1.796 [1.796, 1.823] & 57.40 [57.36, 58.36] & 58.60 [58.46, 58.78] & 2.808 [2.807, 2.815] \\
MultiviewX & GM-PHD & CalibTrack & 83.90 [82.42, 83.90] & 84.85 [84.78, 85.02] & 1.846 [1.833, 1.851] & 56.43 [56.43, 57.82] & 59.43 [59.32, 59.77] & 2.800 [2.791, 2.800] \\
MultiviewX & pruned $\delta$-GLMB & Original ModTrack & 83.62 [83.60, 83.62] & 85.01 [84.94, 85.01] & 1.963 [1.963, 1.965] & 64.80 [64.80, 64.86] & 63.48 [63.48, 63.59] & 2.803 [2.801, 2.804] \\
MultiviewX & pruned $\delta$-GLMB & Confidence-scaled & 80.64 [80.34, 80.69] & 76.24 [75.57, 76.44] & 2.349 [2.349, 2.368] & 64.46 [64.45, 64.66] & 60.84 [60.75, 61.08] & 2.873 [2.870, 2.877] \\
MultiviewX & pruned $\delta$-GLMB & Pooled $R\times T$ & 84.45 [84.45, 84.45] & 83.87 [83.87, 83.87] & 1.959 [1.959, 1.959] & 65.35 [65.35, 65.37] & 62.78 [62.78, 62.82] & 2.788 [2.784, 2.788] \\
MultiviewX & pruned $\delta$-GLMB & CalibTrack & 83.79 [83.70, 83.79] & 85.42 [85.17, 85.49] & 1.935 [1.933, 1.945] & 66.08 [65.97, 66.08] & 63.60 [63.29, 63.60] & 2.797 [2.797, 2.805] \\
MultiviewX & JIPDA & Original ModTrack & 85.74 [85.74, 85.74] & 85.07 [85.07, 85.07] & 2.007 [2.007, 2.007] & 63.49 [63.49, 63.74] & 64.76 [64.65, 64.76] & 2.745 [2.745, 2.746] \\
MultiviewX & JIPDA & Confidence-scaled & 87.43 [87.43, 87.87] & 87.88 [87.55, 87.88] & 1.900 [1.900, 1.906] & 64.66 [64.66, 64.69] & 64.13 [63.96, 64.13] & 2.781 [2.781, 2.782] \\
MultiviewX & JIPDA & Pooled $R\times T$ & 86.79 [86.53, 87.24] & 86.41 [85.41, 86.41] & 1.919 [1.919, 1.932] & 65.21 [65.13, 65.31] & 64.71 [64.68, 65.03] & 2.746 [2.728, 2.746] \\
MultiviewX & JIPDA & CalibTrack & 84.96 [84.90, 84.96] & 86.82 [86.68, 86.82] & 1.919 [1.919, 1.919] & 64.14 [64.09, 64.14] & 64.38 [64.26, 64.38] & 2.746 [2.746, 2.748] \\
\bottomrule
\end{tabular}
}
}
\end{table*}


\clearpage
\section{Evaluation Protocols}
\label{sec:statistical_supplement}

\subsection{Paired Testing Protocol}
\label{sec:paired_protocol_supp}

The paired comparisons use the same detections, confidences, depths, semantic embeddings, BEV measurements, ground truth, test frames, and tracker settings for CalibTrack and each baseline, changing only the covariance. Evaluations use seed 42 and gates $(0.01,0.01)$. IDF1 and MOTA differences are CalibTrack minus baseline, in percentage points; GOSPA benefit is baseline minus CalibTrack. Positive values favor CalibTrack. Standard deviations describe variation across training seeds or temporal-fold means, as specified in each table.

\subsection{Temporal Cross-Validation}
\label{sec:temporal}
Temporal cross-validation pools clean and all six adverse conditions for covariance training, as in the main-paper experiment, while varying the training, validation, and test frame ranges. Each 400-frame sequence is divided into five consecutive 80-frame folds, and each fold is used once for testing. The preceding fold is used for validation and the remaining three folds are used for training. This gives 240 training frames, 80 validation frames, and 80 test frames per condition. The same frame ranges apply to every condition, so test frames are excluded from training and validation in all their clean and adverse versions. We train five covariance heads per fold with different random initializations (seeds 42--46), giving 25 trained heads per dataset. Each head is evaluated on clean and adverse images in its test fold. This tests how the findings depend on the choice of test interval.

\subsection{Leave-One-Condition-Out Validation}
\label{sec:loco}
For leave-one-condition-out (LOCO), we omit one of the six adverse conditions from both covariance training and validation, pooling clean data and the other five adverse conditions. For example, when rain is omitted, the covariance heads are trained on clean, snow, fog, dust, Gaussian-noise, and browning data. For each omitted condition, we train five heads per dataset with five different random initializations (seeds 42--46), using frames 0--279 for training and 280--359 for validation. Each head is then tested on frames 360--399 under the omitted adverse condition. The detector, depth model, and semantic model remain matched to the test condition. For example, when rain is omitted from training CalibTrack, it is used to train the perception modules. LOCO therefore evaluates transfer of covariance estimation while keeping condition-matched perception.

\section{Complete Tracking Results and Adaptive Filtering}
\label{sec:complete_tracking_supp}

Tables~\ref{tab:full_wild} and~\ref{tab:full_mvx} report complete per-condition tracking results for WildTrack and MultiviewX, respectively, including the adaptive-filter baselines described in Section~\ref{sec:control}.

\subsection{Adaptive Filtering}
\label{sec:control}
To compare CalibTrack's learned covariance, which enters the tracker as an input, to a method that instead estimates covariance within the tracker, we evaluate an adaptive-Kalman-style approach~\cite{mehra1970identification,mehra1972approaches,akhlaghi2017adaptive,zhang2020fifty}.
This is not a reproduction of the full adaptive EKF~\cite{akhlaghi2017adaptive}, which estimates both $Q$ and $R$ (process noise and measurement covariance) and uses a residual-based update for $R$ while maintaining positive-definiteness.
Our diagnostic deliberately adapts only $R$, because this control tests whether filter-side measurement-covariance refinement alone can substitute for detector-side covariance prediction.
After associating a detection with a track, we compute the innovation
$\nu_t=\vz_t-H\vx_t^-$, which is the difference between the measured position
$\vz_t$ and the predicted track state position $H\vx_t^-$. We then estimate the measurement-noise covariance from the current
innovation as $\tilde{R}_t=\nu_t\nu_t^\top-HP_t^-H^\top$, subtracting the predicted position uncertainty $HP_t^-H^\top$ from the innovation covariance estimated from the current observation $\nu_t\nu_t^\top$. $P_t^-$ is the predicted track state covariance before incorporating the detection.
We then smooth it using
$R_t^{\mathrm{AKF}}=(1-\beta)R_{t-1}^{\mathrm{AKF}}+\beta\tilde{R}_t$ and then symmetrize.
We use $R_t^{\mathrm{AKF}}$ only in the Kalman update. Association
and cross-view fusion keep the original detector-side covariance.

For each dataset and AKF baseline, we select the smoothing coefficient
$\beta\in\{0.01,0.05,0.10\}$ by maximizing clean-condition IDF1. This selects $\beta=0.10$ for both Original ModTrack and Confidence-scaled on WildTrack,
and $\beta=0.01$ for both baselines on MultiviewX. Both association gates are fixed at
$\alpha_{\mathrm{trk}}=\alpha_{\mathrm{xv}}=0.01$ throughout the sweep and subsequent evaluations.
We keep these hyperparameters fixed across adverse conditions for evaluation.

\begin{table*}[!tbp]
{\centering
\caption{\textbf{Complete Tracking Results on WildTrack.} Reports the per-condition results averaged in main paper Table~\ref{tab:main} along with adaptive filtering results. CalibTrack entries are mean $\pm$ standard deviation across ten trained heads. Avg. adverse averages the six non-clean conditions.}
\label{tab:full_wild}
{\tiny
\setlength{\tabcolsep}{1pt}
\resizebox{\textwidth}{!}{%
\begin{tabular}{lccc|ccc|ccc|ccc}\toprule
 & \multicolumn{3}{c}{Clean} & \multicolumn{3}{c}{Rain} & \multicolumn{3}{c}{Snow} & \multicolumn{3}{c}{Fog} \\
Variant & IDF1 $\uparrow$ & MOTA $\uparrow$ & GOSPA $\downarrow$ & IDF1 $\uparrow$ & MOTA $\uparrow$ & GOSPA $\downarrow$ & IDF1 $\uparrow$ & MOTA $\uparrow$ & GOSPA $\downarrow$ & IDF1 $\uparrow$ & MOTA $\uparrow$ & GOSPA $\downarrow$ \\
\midrule
GM-PHD + Original ModTrack & 95.50 & 91.39 & 1.200 & 68.84 & 67.86 & 1.899 & 76.30 & 76.89 & 1.620 & 72.66 & 73.11 & 1.809 \\
GM-PHD + Confidence-scaled & 84.38 & 86.55 & 1.331 & 70.64 & 65.44 & 1.986 & 77.21 & 75.74 & 1.673 & 71.78 & 70.80 & 1.873 \\
GM-PHD + Pooled $R\times T$ & 91.16 & 89.29 & 1.216 & 67.66 & 66.70 & 1.974 & 76.43 & 76.26 & 1.619 & 75.43 & 71.74 & 1.838 \\
GM-PHD + CalibTrack & \pmcell{89.62}{2.06} & \pmcell{88.49}{0.98} & \pmcell{1.240}{0.055} & \pmcell{69.32}{1.62} & \pmcell{68.00}{1.40} & \pmcell{1.943}{0.035} & \pmcell{78.51}{1.97} & \pmcell{78.11}{0.69} & \pmcell{1.609}{0.020} & \pmcell{77.92}{1.74} & \pmcell{73.87}{1.17} & \pmcell{1.805}{0.040} \\
GM-PHD + Original ModTrack + AKF ($\beta=0.1$) & 94.59 & 89.50 & 1.296 & 67.89 & 63.87 & 2.044 & 75.58 & 77.73 & 1.656 & 80.25 & 74.16 & 1.836 \\
GM-PHD + Confidence-scaled + AKF ($\beta=0.1$) & 86.75 & 82.67 & 1.581 & 70.22 & 65.13 & 2.016 & 78.87 & 76.68 & 1.707 & 75.42 & 73.00 & 1.890 \\
\midrule
pruned $\delta$-GLMB + Original ModTrack & 86.51 & 80.99 & 1.596 & 62.19 & 48.63 & 2.442 & 68.33 & 63.76 & 2.047 & 65.58 & 61.45 & 2.157 \\
pruned $\delta$-GLMB + Confidence-scaled & 77.67 & 71.43 & 1.848 & 52.20 & 26.16 & 2.957 & 65.34 & 51.26 & 2.413 & 61.35 & 48.84 & 2.434 \\
pruned $\delta$-GLMB + Pooled $R\times T$ & 76.47 & 70.69 & 1.852 & 53.47 & 19.54 & 3.076 & 61.75 & 48.32 & 2.487 & 58.17 & 45.27 & 2.501 \\
pruned $\delta$-GLMB + CalibTrack & \pmcell{81.38}{2.39} & \pmcell{75.67}{1.70} & \pmcell{1.748}{0.061} & \pmcell{64.42}{1.56} & \pmcell{43.13}{1.84} & \pmcell{2.667}{0.035} & \pmcell{69.19}{1.68} & \pmcell{55.59}{1.11} & \pmcell{2.333}{0.031} & \pmcell{68.56}{1.43} & \pmcell{57.04}{1.55} & \pmcell{2.334}{0.036} \\
\midrule
JIPDA + Original ModTrack & 90.08 & 86.97 & 1.505 & 66.00 & 55.04 & 2.321 & 74.79 & 70.06 & 1.808 & 69.37 & 67.86 & 1.993 \\
JIPDA + Confidence-scaled & 85.77 & 82.25 & 1.595 & 63.28 & 52.31 & 2.345 & 76.97 & 68.59 & 1.843 & 75.31 & 66.49 & 2.022 \\
JIPDA + Pooled $R\times T$ & 85.84 & 85.40 & 1.554 & 57.37 & 53.47 & 2.330 & 71.06 & 68.80 & 1.839 & 71.54 & 67.54 & 2.024 \\
JIPDA + CalibTrack & \pmcell{89.20}{1.02} & \pmcell{85.79}{0.43} & \pmcell{1.546}{0.055} & \pmcell{69.68}{2.37} & \pmcell{54.92}{1.61} & \pmcell{2.370}{0.021} & \pmcell{77.14}{1.34} & \pmcell{72.09}{1.12} & \pmcell{1.858}{0.028} & \pmcell{77.31}{1.50} & \pmcell{69.47}{0.85} & \pmcell{1.992}{0.024} \\
 & \multicolumn{3}{c}{Dust} & \multicolumn{3}{c}{Gauss $\sigma{=}0.05$} & \multicolumn{3}{c}{Browning} & \multicolumn{3}{c}{Avg. adverse} \\
Variant & IDF1 $\uparrow$ & MOTA $\uparrow$ & GOSPA $\downarrow$ & IDF1 $\uparrow$ & MOTA $\uparrow$ & GOSPA $\downarrow$ & IDF1 $\uparrow$ & MOTA $\uparrow$ & GOSPA $\downarrow$ & IDF1 $\uparrow$ & MOTA $\uparrow$ & GOSPA $\downarrow$ \\
\midrule
GM-PHD + Original ModTrack & 70.59 & 71.43 & 1.784 & 78.26 & 72.48 & 1.848 & 88.22 & 83.82 & 1.491 & 75.81 & 74.26 & 1.742 \\
GM-PHD + Confidence-scaled & 69.87 & 69.12 & 1.887 & 77.07 & 71.11 & 1.911 & 86.08 & 81.83 & 1.550 & 75.44 & 72.34 & 1.813 \\
GM-PHD + Pooled $R\times T$ & 69.32 & 67.86 & 1.883 & 71.86 & 68.28 & 1.918 & 78.48 & 78.78 & 1.566 & 73.20 & 71.60 & 1.800 \\
GM-PHD + CalibTrack & \pmcell{75.76}{1.74} & \pmcell{73.63}{1.10} & \pmcell{1.773}{0.019} & \pmcell{77.84}{1.00} & \pmcell{72.18}{0.96} & \pmcell{1.815}{0.030} & \pmcell{86.19}{2.26} & \pmcell{84.32}{0.38} & \pmcell{1.435}{0.018} & \pmcell{77.59}{0.66} & \pmcell{75.02}{0.38} & \pmcell{1.730}{0.018} \\
GM-PHD + Original ModTrack + AKF ($\beta=0.1$) & 66.37 & 72.90 & 1.842 & 79.91 & 71.53 & 1.926 & 77.77 & 80.25 & 1.627 & 74.63 & 73.41 & 1.822 \\
GM-PHD + Confidence-scaled + AKF ($\beta=0.1$) & 72.78 & 68.17 & 1.946 & 76.62 & 69.85 & 1.994 & 86.17 & 82.56 & 1.573 & 76.68 & 72.57 & 1.854 \\
\midrule
pruned $\delta$-GLMB + Original ModTrack & 59.62 & 50.42 & 2.474 & 63.34 & 54.20 & 2.304 & 73.02 & 67.12 & 1.974 & 65.35 & 57.60 & 2.233 \\
pruned $\delta$-GLMB + Confidence-scaled & 63.34 & 40.02 & 2.684 & 62.87 & 40.13 & 2.699 & 66.82 & 55.57 & 2.325 & 61.99 & 43.66 & 2.585 \\
pruned $\delta$-GLMB + Pooled $R\times T$ & 55.38 & 31.41 & 2.857 & 64.30 & 40.76 & 2.704 & 69.20 & 53.05 & 2.370 & 60.38 & 39.72 & 2.666 \\
pruned $\delta$-GLMB + CalibTrack & \pmcell{62.78}{2.00} & \pmcell{45.82}{1.52} & \pmcell{2.573}{0.030} & \pmcell{66.66}{1.45} & \pmcell{47.69}{1.83} & \pmcell{2.528}{0.040} & \pmcell{73.89}{2.18} & \pmcell{59.68}{2.35} & \pmcell{2.241}{0.048} & \pmcell{67.58}{0.65} & \pmcell{51.49}{0.63} & \pmcell{2.446}{0.018} \\
\midrule
JIPDA + Original ModTrack & 69.31 & 62.08 & 2.170 & 71.00 & 63.13 & 2.067 & 82.08 & 75.95 & 1.792 & 72.09 & 65.69 & 2.025 \\
JIPDA + Confidence-scaled & 69.72 & 60.40 & 2.206 & 71.11 & 61.03 & 2.108 & 79.52 & 74.26 & 1.793 & 72.65 & 63.85 & 2.053 \\
JIPDA + Pooled $R\times T$ & 66.51 & 58.40 & 2.180 & 69.97 & 63.97 & 2.074 & 78.58 & 76.16 & 1.807 & 69.17 & 64.72 & 2.042 \\
JIPDA + CalibTrack & \pmcell{69.18}{3.18} & \pmcell{62.55}{1.26} & \pmcell{2.149}{0.022} & \pmcell{76.19}{1.24} & \pmcell{64.66}{1.55} & \pmcell{2.060}{0.014} & \pmcell{83.57}{1.55} & \pmcell{77.27}{1.06} & \pmcell{1.785}{0.021} & \pmcell{75.51}{0.99} & \pmcell{66.83}{0.47} & \pmcell{2.036}{0.008} \\
\bottomrule
\end{tabular}
}
}
}
\end{table*}

\begin{table*}[!tbp]
\centering
\caption{{\textbf{Complete Tracking Results on MultiviewX.} Same details as Table~\ref{tab:full_wild}.}}
\label{tab:full_mvx}
{\tiny
\setlength{\tabcolsep}{1pt}
\resizebox{\textwidth}{!}{%
\begin{tabular}{lccc|ccc|ccc|ccc}\toprule
 & \multicolumn{3}{c}{Clean} & \multicolumn{3}{c}{Rain} & \multicolumn{3}{c}{Snow} & \multicolumn{3}{c}{Fog} \\
Variant & IDF1 $\uparrow$ & MOTA $\uparrow$ & GOSPA $\downarrow$ & IDF1 $\uparrow$ & MOTA $\uparrow$ & GOSPA $\downarrow$ & IDF1 $\uparrow$ & MOTA $\uparrow$ & GOSPA $\downarrow$ & IDF1 $\uparrow$ & MOTA $\uparrow$ & GOSPA $\downarrow$ \\
\midrule
GM-PHD + Original ModTrack & 86.15 & 86.35 & 1.825 & 48.29 & 48.86 & 3.194 & 51.71 & 55.49 & 2.963 & 57.11 & 62.32 & 2.776 \\
GM-PHD + Confidence-scaled & 82.61 & 84.20 & 1.947 & 49.00 & 47.32 & 3.173 & 54.47 & 56.89 & 2.949 & 54.28 & 61.85 & 2.763 \\
GM-PHD + Pooled $R\times T$ & 82.29 & 84.67 & 1.796 & 50.62 & 46.92 & 3.205 & 52.37 & 54.02 & 2.940 & 55.29 & 61.24 & 2.755 \\
GM-PHD + CalibTrack & \pmcell{83.90}{1.30} & \pmcell{84.85}{0.60} & \pmcell{1.846}{0.060} & \pmcell{47.00}{1.23} & \pmcell{48.39}{0.54} & \pmcell{3.155}{0.009} & \pmcell{53.67}{1.29} & \pmcell{56.39}{0.44} & \pmcell{2.909}{0.017} & \pmcell{55.13}{1.76} & \pmcell{61.33}{1.01} & \pmcell{2.767}{0.024} \\
GM-PHD + Original ModTrack + AKF ($\beta=0.01$) & 84.54 & 85.88 & 1.853 & 47.46 & 48.39 & 3.225 & 49.32 & 55.49 & 2.947 & 54.35 & 60.71 & 2.812 \\
GM-PHD + Confidence-scaled + AKF ($\beta=0.01$) & 78.10 & 84.07 & 1.929 & 50.13 & 49.53 & 3.147 & 54.99 & 55.35 & 2.981 & 57.56 & 61.31 & 2.781 \\
\midrule
pruned $\delta$-GLMB + Original ModTrack & 83.62 & 85.01 & 1.963 & 57.53 & 54.28 & 3.145 & 65.58 & 63.99 & 2.790 & 66.98 & 65.66 & 2.753 \\
pruned $\delta$-GLMB + Confidence-scaled & 80.64 & 76.24 & 2.349 & 58.98 & 53.08 & 3.123 & 66.37 & 60.58 & 2.906 & 65.59 & 62.99 & 2.839 \\
pruned $\delta$-GLMB + Pooled $R\times T$ & 84.45 & 83.87 & 1.959 & 56.47 & 53.15 & 3.138 & 63.91 & 60.11 & 2.833 & 68.38 & 62.25 & 2.858 \\
pruned $\delta$-GLMB + CalibTrack & \pmcell{83.79}{1.55} & \pmcell{85.42}{1.30} & \pmcell{1.935}{0.071} & \pmcell{60.19}{1.19} & \pmcell{55.47}{1.83} & \pmcell{3.070}{0.053} & \pmcell{63.82}{0.75} & \pmcell{63.07}{1.16} & \pmcell{2.807}{0.042} & \pmcell{67.39}{1.50} & \pmcell{65.38}{1.44} & \pmcell{2.759}{0.044} \\
\midrule
JIPDA + Original ModTrack & 85.74 & 85.07 & 2.007 & 55.26 & 54.62 & 3.060 & 61.78 & 62.99 & 2.790 & 64.56 & 68.74 & 2.673 \\
JIPDA + Confidence-scaled & 87.43 & 87.88 & 1.900 & 57.05 & 55.49 & 3.044 & 65.01 & 63.86 & 2.819 & 68.75 & 68.34 & 2.688 \\
JIPDA + Pooled $R\times T$ & 86.79 & 86.41 & 1.919 & 58.60 & 55.62 & 3.047 & 61.62 & 63.32 & 2.763 & 65.77 & 67.87 & 2.667 \\
JIPDA + CalibTrack & \pmcell{84.96}{0.97} & \pmcell{86.82}{0.46} & \pmcell{1.919}{0.026} & \pmcell{57.74}{0.65} & \pmcell{55.32}{0.80} & \pmcell{3.010}{0.020} & \pmcell{61.53}{1.19} & \pmcell{62.26}{0.45} & \pmcell{2.788}{0.014} & \pmcell{65.24}{1.02} & \pmcell{67.99}{0.48} & \pmcell{2.659}{0.019} \\
 & \multicolumn{3}{c}{Dust} & \multicolumn{3}{c}{Gauss $\sigma{=}0.05$} & \multicolumn{3}{c}{Browning} & \multicolumn{3}{c}{Avg. adverse} \\
Variant & IDF1 $\uparrow$ & MOTA $\uparrow$ & GOSPA $\downarrow$ & IDF1 $\uparrow$ & MOTA $\uparrow$ & GOSPA $\downarrow$ & IDF1 $\uparrow$ & MOTA $\uparrow$ & GOSPA $\downarrow$ & IDF1 $\uparrow$ & MOTA $\uparrow$ & GOSPA $\downarrow$ \\
\midrule
GM-PHD + Original ModTrack & 48.58 & 52.41 & 3.055 & 53.88 & 51.94 & 3.168 & 81.08 & 86.61 & 1.826 & 56.77 & 59.61 & 2.830 \\
GM-PHD + Confidence-scaled & 50.65 & 52.61 & 3.057 & 53.07 & 48.73 & 3.247 & 77.50 & 84.87 & 1.887 & 56.49 & 58.71 & 2.846 \\
GM-PHD + Pooled $R\times T$ & 52.38 & 52.74 & 3.030 & 56.00 & 51.74 & 3.107 & 77.72 & 84.94 & 1.812 & 57.40 & 58.60 & 2.808 \\
GM-PHD + CalibTrack & \pmcell{50.85}{1.44} & \pmcell{52.26}{0.50} & \pmcell{3.022}{0.012} & \pmcell{54.39}{2.02} & \pmcell{52.42}{0.38} & \pmcell{3.119}{0.016} & \pmcell{77.51}{1.94} & \pmcell{85.79}{0.99} & \pmcell{1.829}{0.068} & \pmcell{56.43}{0.53} & \pmcell{59.43}{0.34} & \pmcell{2.800}{0.018} \\
GM-PHD + Original ModTrack + AKF ($\beta=0.01$) & 49.00 & 53.21 & 3.050 & 53.23 & 50.67 & 3.224 & 79.55 & 86.01 & 1.838 & 55.48 & 59.08 & 2.849 \\
GM-PHD + Confidence-scaled + AKF ($\beta=0.01$) & 52.36 & 52.41 & 3.064 & 56.51 & 48.80 & 3.263 & 79.02 & 84.20 & 1.934 & 58.43 & 58.60 & 2.861 \\
\midrule
pruned $\delta$-GLMB + Original ModTrack & 56.29 & 55.09 & 3.039 & 61.65 & 56.89 & 3.066 & 80.77 & 84.94 & 2.024 & 64.80 & 63.48 & 2.803 \\
pruned $\delta$-GLMB + Confidence-scaled & 56.93 & 54.02 & 3.103 & 59.25 & 50.33 & 3.267 & 79.61 & 84.07 & 2.001 & 64.46 & 60.84 & 2.873 \\
pruned $\delta$-GLMB + Pooled $R\times T$ & 59.81 & 57.70 & 2.970 & 64.24 & 60.58 & 2.928 & 79.27 & 82.93 & 2.000 & 65.35 & 62.78 & 2.788 \\
pruned $\delta$-GLMB + CalibTrack & \pmcell{58.73}{0.71} & \pmcell{56.10}{1.21} & \pmcell{3.021}{0.041} & \pmcell{64.23}{0.81} & \pmcell{57.72}{1.42} & \pmcell{3.071}{0.045} & \pmcell{82.10}{2.15} & \pmcell{83.83}{3.15} & \pmcell{2.056}{0.130} & \pmcell{66.08}{0.35} & \pmcell{63.60}{0.48} & \pmcell{2.797}{0.026} \\
\midrule
JIPDA + Original ModTrack & 57.02 & 58.37 & 2.959 & 60.41 & 56.02 & 3.063 & 81.91 & 87.82 & 1.926 & 63.49 & 64.76 & 2.745 \\
JIPDA + Confidence-scaled & 56.65 & 56.96 & 3.028 & 57.66 & 53.82 & 3.143 & 82.86 & 86.35 & 1.963 & 64.66 & 64.13 & 2.781 \\
JIPDA + Pooled $R\times T$ & 59.39 & 59.24 & 2.952 & 63.76 & 57.63 & 3.012 & 82.15 & 84.61 & 2.035 & 65.21 & 64.71 & 2.746 \\
JIPDA + CalibTrack & \pmcell{58.66}{0.50} & \pmcell{58.82}{0.56} & \pmcell{2.950}{0.019} & \pmcell{60.31}{0.85} & \pmcell{55.99}{0.64} & \pmcell{3.074}{0.024} & \pmcell{81.39}{0.98} & \pmcell{85.88}{0.78} & \pmcell{1.996}{0.047} & \pmcell{64.14}{0.51} & \pmcell{64.38}{0.21} & \pmcell{2.746}{0.014} \\
\bottomrule
\end{tabular}
}
}
\end{table*}

\section{Measurement Calibration}
\label{sec:measurement_calibration_supp}



Table~\ref{tab:calibration_noise_full} reports empirical calibration metrics on the test set of each dataset. The per-dataset averages give equal weight to all seven conditions, and the cross-dataset average gives equal weight to both datasets.

\begin{table*}[!tbp]
\centering
\scriptsize
\caption{\textbf{Per-Condition Measurement-level calibration.}
CalibTrack averages across ten training seeds. Bold and underline mark the best and second-best estimates by proximity
to the calibration targets.}
\label{tab:calibration_noise_full}
\setlength{\tabcolsep}{3pt}
\renewcommand{\arraystretch}{1.08}

\begin{minipage}[t]{0.49\textwidth}
\centering
\textbf{(a) NEES (target: $1$)}\par
\vspace{4pt}
\begin{adjustbox}{max width=\linewidth}
\begin{tabular}{lrrrrr}
\toprule
Condition
& \shortstack{Original\\ModTrack}
& \shortstack{Confidence-\\scaled}
& \shortstack{Pooled\\$R\times T$}
& $R\times T_c$
& CalibTrack \\
\midrule
\multicolumn{6}{l}{\textbf{WildTrack}} \\
\addlinespace[2pt]
Clean & 0.451 & \second{1.465} & 4.003 & 1.614 & \textbf{1.127} \\
Rain & \textbf{1.221} & 3.906 & 10.845 & \second{1.625} & 2.264 \\
Snow & \textbf{0.863} & 2.713 & 7.666 & 1.803 & \second{1.706} \\
Fog & \textbf{0.824} & 2.664 & 7.321 & 1.713 & \second{1.519} \\
Dust & \textbf{0.961} & 2.998 & 8.533 & \second{1.478} & 1.626 \\
Gauss $\sigma{=}0.05$
 & \textbf{0.821} & 2.548 & 7.290 & 1.690 & \second{1.297} \\
Browning & \second{0.490} & 1.521 & 4.349 & 2.119 & \textbf{1.082} \\
\midrule
\textit{Average}
 & \textbf{0.804} & 2.545 & 7.144 & 1.720 & \second{1.517} \\
\midrule
\multicolumn{6}{l}{\textbf{MultiviewX}} \\
\addlinespace[2pt]
Clean & 0.370 & \textbf{0.911} & 2.026 & \second{1.152} & 0.804 \\
Rain & \second{0.750} & 1.647 & 4.105 & \textbf{1.112} & 1.269 \\
Snow & 0.662 & 1.574 & 3.623 & \second{1.219} & \textbf{1.109} \\
Fog & 0.633 & 1.365 & 3.461 & \second{1.170} & \textbf{1.100} \\
Dust & 0.606 & 1.275 & 3.316 & \second{1.110} & \textbf{1.022} \\
Gauss $\sigma{=}0.05$
 & 0.549 & 1.345 & 3.004 & \textbf{1.048} & \second{0.925} \\
Browning & 0.355 & \textbf{0.852} & 1.942 & \second{1.168} & 0.775 \\
\midrule
\textit{Average}
 & 0.561 & 1.281 & 3.068 & \second{1.140} & \textbf{1.001} \\
\midrule
\multicolumn{6}{l}{\textbf{Cross dataset}} \\
\addlinespace[2pt]
\textit{Average}
 & \second{0.683} & 1.913 & 5.106 & 1.430 & \textbf{1.259} \\
\bottomrule
\end{tabular}
\end{adjustbox}
\end{minipage}
\hfill
\begin{minipage}[t]{0.49\textwidth}
\centering
\textbf{(b) $2\sigma$ coverage (target: $86.47\%$)}\par
\vspace{4pt}
\begin{adjustbox}{max width=\linewidth}
\begin{tabular}{lrrrrr}
\toprule
Condition
& \shortstack{Original\\ModTrack}
& \shortstack{Confidence-\\scaled}
& \shortstack{Pooled\\$R\times T$}
& $R\times T_c$
& CalibTrack \\
\midrule
\multicolumn{6}{l}{\textbf{WildTrack}} \\
\addlinespace[2pt]
Clean & 94.42 & 90.87 & 80.34 & \textbf{87.95} & \second{90.02} \\
Rain & \textbf{86.33} & 82.67 & 58.91 & \second{83.75} & 81.05 \\
Snow & 89.93 & \textbf{85.50} & 64.78 & 83.12 & \second{84.58} \\
Fog & 90.03 & \textbf{85.95} & 64.83 & 84.19 & \second{84.29} \\
Dust & 89.00 & \textbf{85.11} & 64.25 & \second{84.90} & 84.80 \\
Gauss $\sigma{=}0.05$
 & 89.65 & \second{85.29} & 67.27 & 83.93 & \textbf{85.49} \\
Browning & 93.64 & 90.99 & 77.00 & \textbf{85.06} & \second{89.99} \\
\midrule
\textit{Average}
 & 90.43 & \textbf{86.63} & 68.20 & 84.70 & \second{85.75} \\
\midrule
\multicolumn{6}{l}{\textbf{MultiviewX}} \\
\addlinespace[2pt]
Clean & 95.55 & 92.53 & \second{83.79} & \textbf{88.50} & 91.47 \\
Rain & 90.03 & \textbf{86.30} & 66.32 & \second{85.86} & 83.38 \\
Snow & 91.55 & 88.42 & 73.27 & \textbf{86.82} & \second{87.69} \\
Fog & 91.90 & 88.70 & 71.40 & \second{86.26} & \textbf{86.42} \\
Dust & 92.44 & 88.82 & 72.48 & \textbf{86.93} & \second{87.71} \\
Gauss $\sigma{=}0.05$
 & 93.82 & 90.06 & 74.55 & \textbf{88.02} & \second{89.79} \\
Browning & 95.93 & 92.92 & \second{83.90} & \textbf{88.10} & 91.64 \\
\midrule
\textit{Average}
 & 93.03 & 89.68 & 75.10 & \textbf{87.21} & \second{88.30} \\
\midrule
\multicolumn{6}{l}{\textbf{Cross dataset}} \\
\addlinespace[2pt]
\textit{Average}
 & 91.73 & 88.15 & 71.65 & \textbf{85.96} & \second{87.02} \\
\bottomrule
\end{tabular}
\end{adjustbox}
\end{minipage}
\end{table*}

\section{Robustness to Camera Removal
\label{sec:camera_dropout_supp}}
We evaluate camera-dropout robustness using the identity-informed GM-PHD core, the same ten covariance heads originally trained on all cameras, test frames 360--399, the established gates $(0.01,0.01)$, and condition-matched perception. Removed cameras remain absent throughout this 40-frame window. For each condition (including avg. adverse), Tables~\ref{tab:camera_dropout}--~\ref{tab:camera_dropout_browning} average the full-window scores equally over all unordered camera subsets and the ten trained heads. WildTrack has 120 subsets ($\sum_{i=0}^5 \binom{7}{i}$) and MultiviewX has 57 subsets ( $\sum_{i=0}^4 \binom{6}{i}$), across $i$ removed cameras. Removing five and six cameras from MultiviewX and WildTrack respectively leaves only one view and is excluded. The zero-removal row uses the matched all-camera reference.

\begin{table*}[!htbp]
{\centering\scriptsize
\caption{\textbf{CalibTrack under camera dropout: Clean.} Entries are mean $\pm$ standard deviation across ten training seeds. Removing five cameras from MultiviewX leaves one view and is excluded.}
\label{tab:camera_dropout}
\setlength{\tabcolsep}{2pt}
\begin{adjustbox}{max width=\textwidth, center=\textwidth}
\begin{tabular}{lccc|ccc}\toprule
Cameras removed & \multicolumn{3}{c|}{WildTrack} & \multicolumn{3}{c}{MultiviewX} \\
 & IDF1 $\uparrow$ & MOTA $\uparrow$ & GOSPA $\downarrow$ & IDF1 $\uparrow$ & MOTA $\uparrow$ & GOSPA $\downarrow$ \\

0 & \pmcell{89.62}{2.06} & \pmcell{88.49}{0.98} & \pmcell{1.240}{0.055} & \pmcell{83.90}{1.30} & \pmcell{84.85}{0.60} & \pmcell{1.846}{0.060} \\
1 & \pmcell{88.31}{1.16} & \pmcell{86.11}{0.64} & \pmcell{1.328}{0.039} & \pmcell{78.35}{0.48} & \pmcell{80.51}{0.50} & \pmcell{2.064}{0.043} \\
2 & \pmcell{83.97}{0.71} & \pmcell{80.31}{0.36} & \pmcell{1.512}{0.025} & \pmcell{71.72}{0.38} & \pmcell{72.82}{0.33} & \pmcell{2.375}{0.029} \\
3 & \pmcell{74.74}{0.30} & \pmcell{69.69}{0.18} & \pmcell{1.832}{0.013} & \pmcell{59.21}{0.26} & \pmcell{57.24}{0.24} & \pmcell{2.884}{0.017} \\
4 & \pmcell{58.11}{0.17} & \pmcell{51.32}{0.10} & \pmcell{2.319}{0.005} & \pmcell{34.62}{0.23} & \pmcell{30.25}{0.12} & \pmcell{3.617}{0.007} \\
5 & \pmcell{31.40}{0.18} & \pmcell{25.22}{0.06} & \pmcell{2.919}{0.001} & --- & --- & --- \\

\bottomrule
\end{tabular}
\end{adjustbox}}
\end{table*}

\begin{table*}[!htbp]
{\centering\scriptsize
\caption{\textbf{CalibTrack under camera dropout: Avg. Adverse.} Entries are mean $\pm$ standard deviation across ten training seeds. Removing five cameras from MultiviewX leaves one view and is excluded.}
\label{tab:camera_dropout_average_adverse}
\setlength{\tabcolsep}{2pt}
\begin{adjustbox}{max width=\textwidth, center=\textwidth}
\begin{tabular}{lccc|ccc}\toprule
Cameras removed & \multicolumn{3}{c|}{WildTrack} & \multicolumn{3}{c}{MultiviewX} \\
 & IDF1 $\uparrow$ & MOTA $\uparrow$ & GOSPA $\downarrow$ & IDF1 $\uparrow$ & MOTA $\uparrow$ & GOSPA $\downarrow$ \\

0 & \pmcell{77.59}{0.66} & \pmcell{75.02}{0.38} & \pmcell{1.730}{0.018} & \pmcell{56.43}{0.53} & \pmcell{59.43}{0.34} & \pmcell{2.800}{0.018} \\
1 & \pmcell{72.91}{0.38} & \pmcell{69.46}{0.20} & \pmcell{1.895}{0.009} & \pmcell{51.78}{0.25} & \pmcell{52.25}{0.16} & \pmcell{3.023}{0.011} \\
2 & \pmcell{65.09}{0.23} & \pmcell{60.22}{0.10} & \pmcell{2.141}{0.005} & \pmcell{43.68}{0.12} & \pmcell{41.69}{0.11} & \pmcell{3.320}{0.008} \\
3 & \pmcell{53.17}{0.09} & \pmcell{46.80}{0.06} & \pmcell{2.460}{0.003} & \pmcell{32.12}{0.09} & \pmcell{27.98}{0.06} & \pmcell{3.678}{0.004} \\
4 & \pmcell{36.76}{0.05} & \pmcell{29.94}{0.05} & \pmcell{2.829}{0.002} & \pmcell{16.21}{0.05} & \pmcell{12.58}{0.05} & \pmcell{4.048}{0.002} \\
5 & \pmcell{17.26}{0.06} & \pmcell{12.59}{0.04} & \pmcell{3.185}{0.001} & --- & --- & --- \\

\bottomrule
\end{tabular}
\end{adjustbox}}
\end{table*}

\begin{table*}[!htbp]
{\centering\scriptsize
\caption{\textbf{CalibTrack under camera dropout: Rain.} Entries are mean $\pm$ standard deviation across ten training seeds. Removing five cameras from MultiviewX leaves one view and is excluded.}
\label{tab:camera_dropout_rain}
\setlength{\tabcolsep}{2pt}
\begin{adjustbox}{max width=\textwidth, center=\textwidth}
\begin{tabular}{lccc|ccc}\toprule
Cameras removed & \multicolumn{3}{c|}{WildTrack} & \multicolumn{3}{c}{MultiviewX} \\
 & IDF1 $\uparrow$ & MOTA $\uparrow$ & GOSPA $\downarrow$ & IDF1 $\uparrow$ & MOTA $\uparrow$ & GOSPA $\downarrow$ \\

0 & \pmcell{69.32}{1.62} & \pmcell{68.00}{1.40} & \pmcell{1.943}{0.035} & \pmcell{47.00}{1.23} & \pmcell{48.39}{0.54} & \pmcell{3.155}{0.009} \\
1 & \pmcell{65.11}{0.74} & \pmcell{62.45}{0.64} & \pmcell{2.099}{0.009} & \pmcell{42.70}{0.51} & \pmcell{41.05}{0.40} & \pmcell{3.376}{0.010} \\
2 & \pmcell{57.08}{0.48} & \pmcell{53.44}{0.22} & \pmcell{2.321}{0.004} & \pmcell{32.85}{0.16} & \pmcell{29.63}{0.22} & \pmcell{3.667}{0.005} \\
3 & \pmcell{44.79}{0.29} & \pmcell{39.71}{0.13} & \pmcell{2.627}{0.004} & \pmcell{21.76}{0.13} & \pmcell{17.33}{0.08} & \pmcell{3.956}{0.001} \\
4 & \pmcell{29.74}{0.31} & \pmcell{23.69}{0.11} & \pmcell{2.962}{0.002} & \pmcell{9.15}{0.13} & \pmcell{6.54}{0.05} & \pmcell{4.191}{0.001} \\
5 & \pmcell{13.28}{0.19} & \pmcell{9.12}{0.07} & \pmcell{3.256}{0.001} & --- & --- & --- \\

\bottomrule
\end{tabular}
\end{adjustbox}}
\end{table*}

\begin{table*}[!tbp]
{\centering\scriptsize
\caption{\textbf{CalibTrack under camera dropout: Snow.} Entries are mean $\pm$ standard deviation across ten training seeds. Removing five cameras from MultiviewX leaves one view and is excluded.}
\label{tab:camera_dropout_snow}
\setlength{\tabcolsep}{2pt}
\begin{adjustbox}{max width=\textwidth, center=\textwidth}
\begin{tabular}{lccc|ccc}\toprule
Cameras removed & \multicolumn{3}{c|}{WildTrack} & \multicolumn{3}{c}{MultiviewX} \\
 & IDF1 $\uparrow$ & MOTA $\uparrow$ & GOSPA $\downarrow$ & IDF1 $\uparrow$ & MOTA $\uparrow$ & GOSPA $\downarrow$ \\

0 & \pmcell{78.51}{1.97} & \pmcell{78.11}{0.69} & \pmcell{1.609}{0.020} & \pmcell{53.67}{1.29} & \pmcell{56.39}{0.44} & \pmcell{2.909}{0.017} \\
1 & \pmcell{75.08}{0.77} & \pmcell{72.15}{0.25} & \pmcell{1.804}{0.009} & \pmcell{48.14}{0.59} & \pmcell{48.24}{0.21} & \pmcell{3.146}{0.006} \\
2 & \pmcell{65.47}{0.47} & \pmcell{61.68}{0.24} & \pmcell{2.088}{0.005} & \pmcell{38.78}{0.27} & \pmcell{36.98}{0.09} & \pmcell{3.455}{0.003} \\
3 & \pmcell{52.15}{0.39} & \pmcell{47.11}{0.19} & \pmcell{2.442}{0.004} & \pmcell{27.02}{0.23} & \pmcell{22.86}{0.08} & \pmcell{3.810}{0.002} \\
4 & \pmcell{35.30}{0.25} & \pmcell{29.63}{0.13} & \pmcell{2.830}{0.002} & \pmcell{12.18}{0.15} & \pmcell{9.03}{0.06} & \pmcell{4.132}{0.002} \\
5 & \pmcell{16.34}{0.17} & \pmcell{12.22}{0.06} & \pmcell{3.190}{0.001} & --- & --- & --- \\

\bottomrule
\end{tabular}
\end{adjustbox}}
\end{table*}

\begin{table*}[!tbp]
{\centering\scriptsize
\caption{\textbf{CalibTrack under camera dropout: Fog.} Entries are mean $\pm$ standard deviation across ten training seeds. Removing five cameras from MultiviewX leaves one view and is excluded.}
\label{tab:camera_dropout_fog}
\setlength{\tabcolsep}{2pt}
\begin{adjustbox}{max width=\textwidth, center=\textwidth}
\begin{tabular}{lccc|ccc}\toprule
Cameras removed & \multicolumn{3}{c|}{WildTrack} & \multicolumn{3}{c}{MultiviewX} \\
 & IDF1 $\uparrow$ & MOTA $\uparrow$ & GOSPA $\downarrow$ & IDF1 $\uparrow$ & MOTA $\uparrow$ & GOSPA $\downarrow$ \\

0 & \pmcell{77.92}{1.74} & \pmcell{73.87}{1.17} & \pmcell{1.805}{0.040} & \pmcell{55.13}{1.76} & \pmcell{61.33}{1.01} & \pmcell{2.767}{0.024} \\
1 & \pmcell{69.33}{0.86} & \pmcell{67.12}{0.57} & \pmcell{1.966}{0.020} & \pmcell{51.45}{0.74} & \pmcell{54.30}{0.40} & \pmcell{2.988}{0.012} \\
2 & \pmcell{61.49}{0.47} & \pmcell{57.95}{0.30} & \pmcell{2.200}{0.009} & \pmcell{44.40}{0.27} & \pmcell{43.06}{0.19} & \pmcell{3.303}{0.007} \\
3 & \pmcell{50.46}{0.30} & \pmcell{44.74}{0.14} & \pmcell{2.512}{0.004} & \pmcell{31.86}{0.17} & \pmcell{27.47}{0.13} & \pmcell{3.702}{0.004} \\
4 & \pmcell{34.97}{0.13} & \pmcell{28.47}{0.12} & \pmcell{2.865}{0.002} & \pmcell{15.39}{0.11} & \pmcell{11.25}{0.07} & \pmcell{4.084}{0.002} \\
5 & \pmcell{16.38}{0.12} & \pmcell{11.93}{0.07} & \pmcell{3.200}{0.001} & --- & --- & --- \\

\bottomrule
\end{tabular}
\end{adjustbox}}
\end{table*}

\begin{table*}[!tbp]
{\centering\scriptsize
\caption{\textbf{CalibTrack under camera dropout: Dust.} Entries are mean $\pm$ standard deviation across ten training seeds. Removing five cameras from MultiviewX leaves one view and is excluded.}
\label{tab:camera_dropout_dust}
\setlength{\tabcolsep}{2pt}
\begin{adjustbox}{max width=\textwidth, center=\textwidth}
\begin{tabular}{lccc|ccc}\toprule
Cameras removed & \multicolumn{3}{c|}{WildTrack} & \multicolumn{3}{c}{MultiviewX} \\
 & IDF1 $\uparrow$ & MOTA $\uparrow$ & GOSPA $\downarrow$ & IDF1 $\uparrow$ & MOTA $\uparrow$ & GOSPA $\downarrow$ \\

0 & \pmcell{75.76}{1.74} & \pmcell{73.63}{1.10} & \pmcell{1.773}{0.019} & \pmcell{50.85}{1.44} & \pmcell{52.26}{0.50} & \pmcell{3.022}{0.012} \\
1 & \pmcell{69.65}{1.00} & \pmcell{67.07}{0.70} & \pmcell{1.977}{0.010} & \pmcell{45.34}{0.65} & \pmcell{45.06}{0.32} & \pmcell{3.252}{0.008} \\
2 & \pmcell{61.87}{0.66} & \pmcell{57.11}{0.27} & \pmcell{2.230}{0.006} & \pmcell{36.95}{0.28} & \pmcell{33.94}{0.25} & \pmcell{3.547}{0.006} \\
3 & \pmcell{50.23}{0.35} & \pmcell{43.69}{0.12} & \pmcell{2.538}{0.003} & \pmcell{25.50}{0.17} & \pmcell{21.05}{0.11} & \pmcell{3.862}{0.003} \\
4 & \pmcell{34.31}{0.21} & \pmcell{27.35}{0.07} & \pmcell{2.887}{0.001} & \pmcell{11.96}{0.14} & \pmcell{8.66}{0.07} & \pmcell{4.143}{0.001} \\
5 & \pmcell{15.86}{0.17} & \pmcell{11.16}{0.05} & \pmcell{3.215}{0.001} & --- & --- & --- \\

\bottomrule
\end{tabular}
\end{adjustbox}}
\end{table*}

\begin{table*}[!tbp]
{\centering\scriptsize
\caption{\textbf{CalibTrack under camera dropout: Gauss Noise $\sigma{=}0.05$.} Entries are mean $\pm$ standard deviation across ten training seeds. Removing five cameras from MultiviewX leaves one view and is excluded.}
\label{tab:camera_dropout_gaussian005}
\setlength{\tabcolsep}{2pt}
\begin{adjustbox}{max width=\textwidth, center=\textwidth}
\begin{tabular}{lccc|ccc}\toprule
Cameras removed & \multicolumn{3}{c|}{WildTrack} & \multicolumn{3}{c}{MultiviewX} \\
 & IDF1 $\uparrow$ & MOTA $\uparrow$ & GOSPA $\downarrow$ & IDF1 $\uparrow$ & MOTA $\uparrow$ & GOSPA $\downarrow$ \\

0 & \pmcell{77.84}{1.00} & \pmcell{72.18}{0.96} & \pmcell{1.815}{0.030} & \pmcell{54.39}{2.02} & \pmcell{52.42}{0.38} & \pmcell{3.119}{0.016} \\
1 & \pmcell{73.98}{0.43} & \pmcell{67.67}{0.46} & \pmcell{1.971}{0.016} & \pmcell{48.96}{0.40} & \pmcell{44.00}{0.29} & \pmcell{3.330}{0.008} \\
2 & \pmcell{65.43}{0.47} & \pmcell{58.11}{0.23} & \pmcell{2.226}{0.008} & \pmcell{40.07}{0.26} & \pmcell{33.89}{0.15} & \pmcell{3.576}{0.005} \\
3 & \pmcell{52.75}{0.29} & \pmcell{44.58}{0.14} & \pmcell{2.535}{0.004} & \pmcell{29.16}{0.20} & \pmcell{22.30}{0.08} & \pmcell{3.850}{0.002} \\
4 & \pmcell{35.15}{0.22} & \pmcell{27.97}{0.11} & \pmcell{2.880}{0.002} & \pmcell{14.41}{0.13} & \pmcell{10.14}{0.07} & \pmcell{4.118}{0.002} \\
5 & \pmcell{15.48}{0.15} & \pmcell{11.43}{0.08} & \pmcell{3.207}{0.001} & --- & --- & --- \\

\bottomrule
\end{tabular}
\end{adjustbox}}
\end{table*}

\begin{table*}[!tbp]
{\centering\scriptsize
\caption{\textbf{CalibTrack under camera dropout: Browning.} Entries are mean $\pm$ standard deviation across ten training seeds. Removing five cameras from MultiviewX leaves one view and is excluded.}
\label{tab:camera_dropout_browning}
\setlength{\tabcolsep}{2pt}
\begin{adjustbox}{max width=\textwidth, center=\textwidth}
\begin{tabular}{lccc|ccc}\toprule
Cameras removed & \multicolumn{3}{c|}{WildTrack} & \multicolumn{3}{c}{MultiviewX} \\
 & IDF1 $\uparrow$ & MOTA $\uparrow$ & GOSPA $\downarrow$ & IDF1 $\uparrow$ & MOTA $\uparrow$ & GOSPA $\downarrow$ \\

0 & \pmcell{86.19}{2.26} & \pmcell{84.32}{0.38} & \pmcell{1.435}{0.018} & \pmcell{77.51}{1.94} & \pmcell{85.79}{0.99} & \pmcell{1.829}{0.068} \\
1 & \pmcell{84.28}{0.90} & \pmcell{80.32}{0.21} & \pmcell{1.556}{0.011} & \pmcell{74.08}{1.03} & \pmcell{80.85}{0.46} & \pmcell{2.043}{0.042} \\
2 & \pmcell{79.23}{0.44} & \pmcell{73.04}{0.22} & \pmcell{1.779}{0.007} & \pmcell{69.01}{0.38} & \pmcell{72.64}{0.29} & \pmcell{2.370}{0.028} \\
3 & \pmcell{68.61}{0.24} & \pmcell{60.96}{0.16} & \pmcell{2.109}{0.004} & \pmcell{57.43}{0.29} & \pmcell{56.85}{0.20} & \pmcell{2.887}{0.017} \\
4 & \pmcell{51.08}{0.18} & \pmcell{42.54}{0.12} & \pmcell{2.552}{0.003} & \pmcell{34.14}{0.25} & \pmcell{29.90}{0.17} & \pmcell{3.621}{0.008} \\
5 & \pmcell{26.23}{0.14} & \pmcell{19.69}{0.06} & \pmcell{3.044}{0.001} & --- & --- & --- \\

\bottomrule
\end{tabular}
\end{adjustbox}}
\end{table*}

\section{Analysis of Heteroscedasticity Gains}
\label{sec:hetero}

\begin{figure*}[!tbp]
    \centering
    \includegraphics[width=\textwidth]{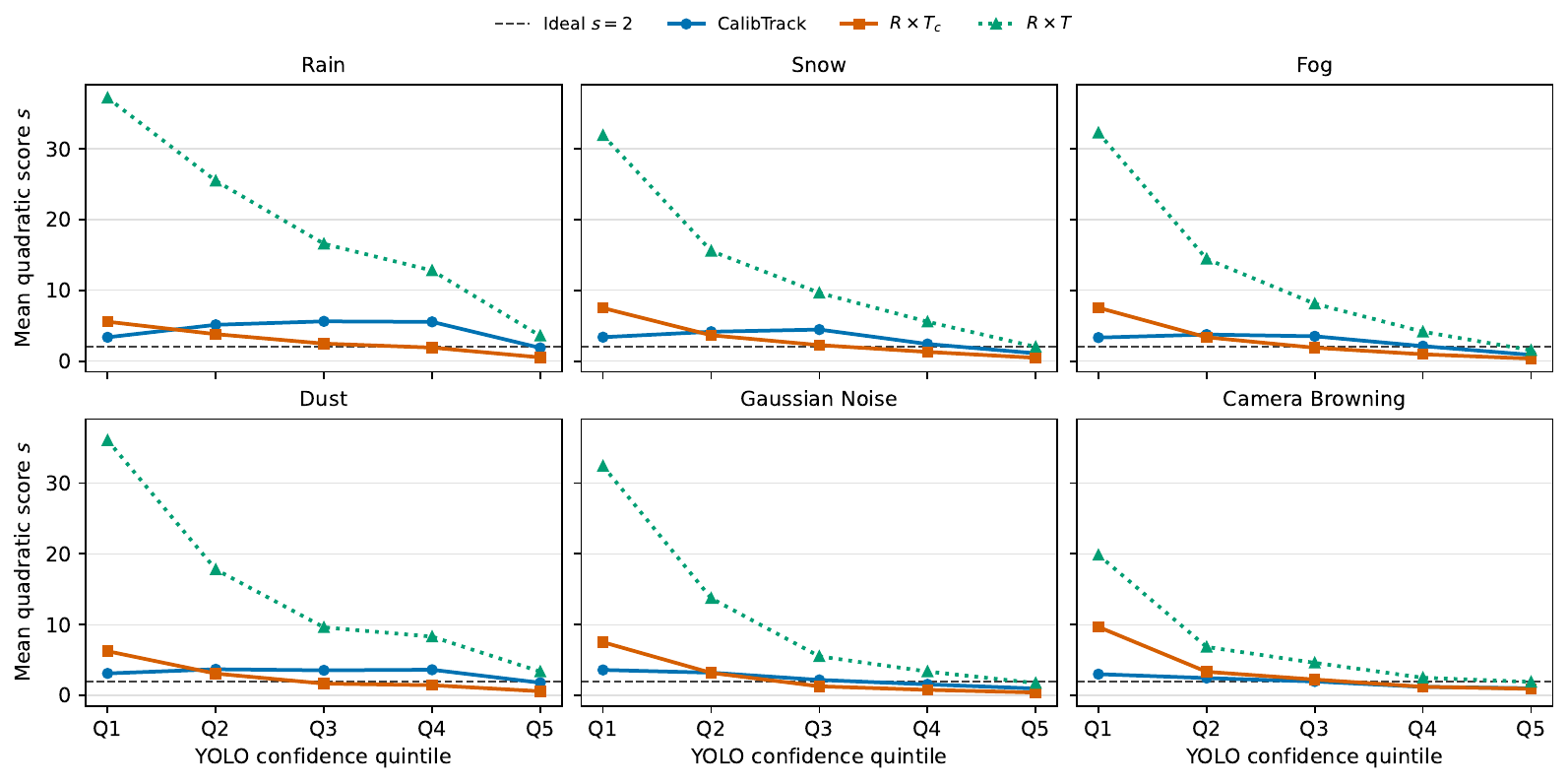}
    \caption{\textbf{Calibration across detection-confidence quintiles on WildTrack.}
    Mean quadratic scores are shown from the lowest (Q1) to highest (Q5) detection confidences. The ideal value is 2 for calibrated 2D Gaussian covariances. Higher values indicate overconfidence and lower values indicate underconfidence.}
    \label{fig:confidence_quintiles_all}
\end{figure*}

\begin{figure*}[!tbp]
    \centering
    \includegraphics[width=\textwidth]{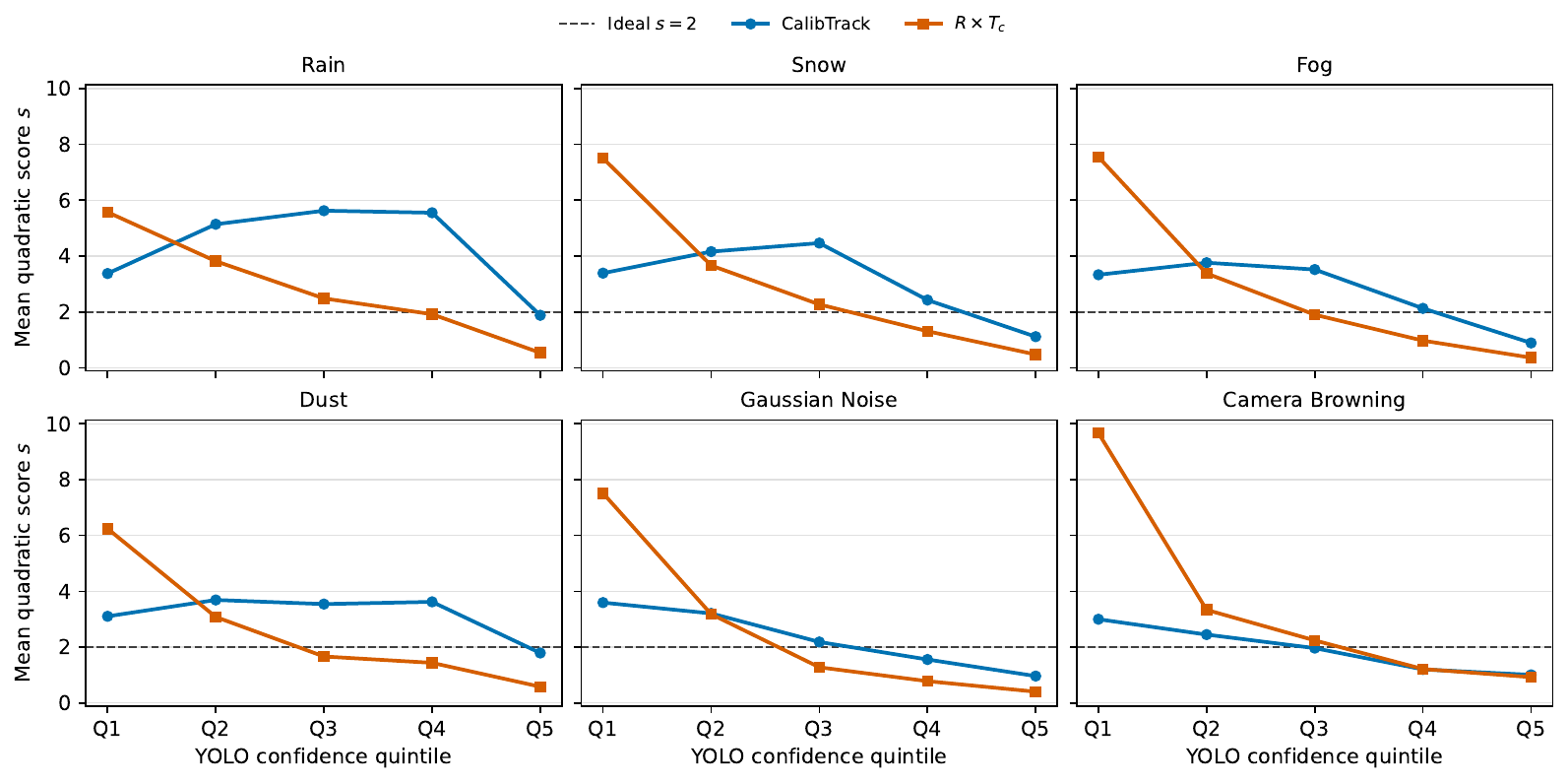}
    \caption{\textbf{Calibration across detection-confidence quintiles on WildTrack excluding $R\times T$.} Same details as Figure~\ref{fig:confidence_quintiles_all} excluding the $R\times T$ baseline.}
    \label{fig:confidence_quintiles_condition_specific}
\end{figure*}

Figures~\ref{fig:confidence_quintiles_all} and ~\ref{fig:confidence_quintiles_condition_specific} show per-adverse-condition calibration results on WildTrack across CalibTrack, $R\times T$, and $R\times T_c$. We include figure~\ref{fig:confidence_quintiles_condition_specific} without $R\times T$ to allow for easier visual comparison of CalibTrack and $R\times T_c$. These figures support the discussion from the main paper that the two scalar baselines are overconfident on the lowest confidence detections and underconfident on the highest confidence detections while CalibTrack remains less sensitive across the confidence quintiles. On rain and dust $R\times T_c$ achieves better calibration at the intermediate confidence levels (Q2--4); however, across all conditions CalibTrack achieves a mean quadratic score closer to the ideal at the tail confidence quintiles  (Q1 and Q5), with $R\times T_c$ consistently overconfident on low confidence detections and underconfident on high confidence detections. We believe this reduced sensitivity to detection confidence enhances CalibTrack's robustness under adverse conditions.

\end{document}